\documentclass[letterpaper, 10 pt, conference]{ieeeconf}
\IEEEoverridecommandlockouts                            

\usepackage[utf8]{inputenc}
\usepackage[T1]{fontenc}
\usepackage{graphicx}
\usepackage{float} 
\usepackage{ragged2e}
\usepackage{booktabs}
\usepackage{bbding}
\usepackage{multirow}
\usepackage{makecell}
\usepackage{amsmath,amssymb,amsfonts}
\usepackage[table,xcdraw]{xcolor}
\usepackage{cite}
\usepackage[colorlinks,linkcolor=blue]{hyperref}
\usepackage{cleveref}
\usepackage{tabularx}
\usepackage[caption=true,font=normalsize,labelfont=sf,textfont=sf]{subfig}
\usepackage{textcomp}
\usepackage{stfloats}
\usepackage{tabularx}
\usepackage[linesnumbered,ruled,vlined]{algorithm2e}
\usepackage{adjustbox} 

\hypersetup{
    colorlinks=true, 
    linkcolor=blue,  
    filecolor=blue,  
    urlcolor=black,  
    citecolor=blue, 
    pdfborder={0 0 0}
}

\Crefname{figure}{Fig.}{Figs.}
\Crefname{equation}{Eq.}{Eqs.}

\def\ie{\emph{i.e.}}

\newcommand{\verticaltext}[3][0pt]{%
  \rotatebox[origin=l]{90}{\vspace{#1} #2 #3}%
}

\title{\bf Label Anything: An Interpretable, High-Fidelity \\and Prompt-Free Annotator
}

\author{~~~Wei-Bin Kou$^{1,2,3}$, Guangxu Zhu$^{3}$, Rongguang Ye$^{2}$, Shuai Wang$^{4}$, Ming Tang$^{2,*}$, and Yik-Chung Wu$^{1,*}$
\thanks{This work has been accepted by 2025 IEEE International Conference on Robotics and Automation (ICRA 2025). This work was supported in part by National Natural Science Foundation of China under Grant 62371313, in part by Shenzhen-Hong Kong-Macau Technology Research Programme (Type C) under Grant SGDX20230821091559018, in part by Hetao Shenzhen-Hong Kong Science and Technology Innovation Cooperation Zone Project under grant HZQSWS-KCCYB-2024016, and in part by the National Natural Science Foundation of China under Grant 62371444).}
\thanks{$^{*}$Corresponding author: Ming Tang (tangm3@sustech.edu.cn) and Yik-Chung Wu (ycwu@eee.hku.hk).}
\thanks{$^{1}$Department of Electrical and Electronic Engineering, The University of Hong Kong, Hong Kong, China.}
\thanks{$^{2}$Department of Computer Science and Engineering, Southern University of Science and Technology, Shenzhen, China.}
\thanks{$^{3}$Shenzhen International Center For Industrial And Applied Mathematics, Shenzhen Research Institute of Big Data, Shenzhen, China.}
\thanks{$^{4}$Shenzhen Institute of Advanced Technology, Chinese Academy of Sciences, Shenzhen, China.}
}

\begin{document}

\maketitle
\thispagestyle{empty}
\pagestyle{empty}

\begin{abstract}
Learning-based street scene semantic understanding in autonomous driving (AD) has advanced significantly recently, but the performance of the AD model is heavily dependent on the quantity and quality of the annotated training data. However, traditional manual labeling involves high cost to annotate the vast amount of required data for training robust model. To mitigate this cost of manual labeling, we propose a \underline{L}abel \underline{A}nything \underline{M}odel (denoted as LAM), serving as an interpretable, high-fidelity, and prompt-free data annotator. Specifically, we firstly incorporate a pretrained Vision Transformer (ViT) to extract the latent features. On top of ViT, we propose a \underline{s}emantic \underline{c}lass \underline{a}dapter (SCA) and an \underline{opt}imization-\underline{o}riented \underline{u}nrolling algorithm (OptOU), both with a quite small number of trainable parameters. SCA is proposed to fuse ViT-extracted features to consolidate the basis of the subsequent automatic annotation. OptOU consists of multiple cascading layers and each layer contains an optimization formulation to align its output with the ground truth as closely as possible, though which OptOU acts as being interpretable rather than learning-based blackbox nature. In addition, training SCA and OptOU requires only a single pre-annotated RGB seed image, owing to their small volume of learnable parameters. Extensive experiments clearly demonstrate that the proposed LAM can generate high-fidelity annotations (almost 100\% in mIoU) for multiple real-world datasets (\ie, Camvid, Cityscapes, and Apolloscapes) and CARLA simulation dataset.
\end{abstract}

\section{INTRODUCTION}
\vspace{-0.2cm}
The task of street scene semantic understanding is pivotal yet intricate for autonomous driving (AD) \cite{10416354,10342110,kou2024fedrc,9815141,kou2024pfedlvmlargevisionmodel,10049523,kou2025enhancing}. Recent advancements have introduced numerous innovative approaches \cite{can2022understanding,kou2024adverse,natan2022towards,kou2024fast}. For instance, \cite{can2022understanding} explores semantic comprehension of roads using onboard Bird's Eye View (BEV) cameras; \cite{natan2022towards} proposes an adaptive loss weighting strategy to address imbalances in learning and integrates multi-sensor fusion to enhance the understanding of dynamic environments. Despite such advancements, these methods are generally data-hungry. In principle, models trained with larger-volume and higher-quality data exhibit improved performance. However, conventional annotation of such volume of training data by human involves a great of time and economic expense. 

\begin{figure*}[tp]
\centering 
\vspace{-0.1cm}
\subfloat[Raw image]{
\label{Fig.row_image}
\includegraphics[width=0.26\linewidth,height=0.12\linewidth]{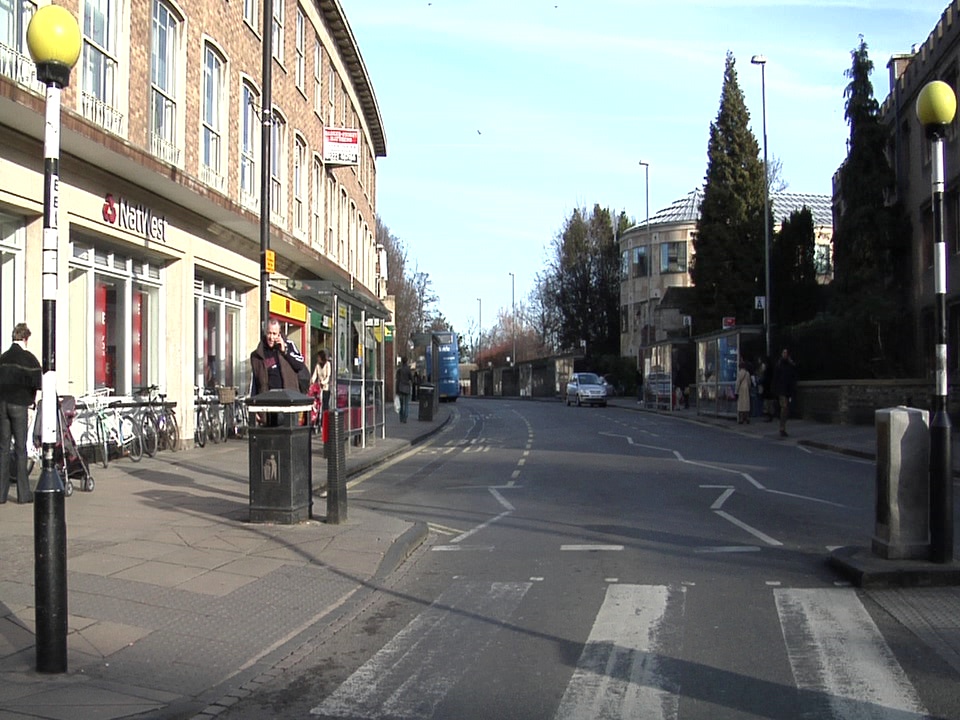}}
\subfloat[SAM's segmentation]{
\hspace{0.3cm}
\label{Fig.SAM_generated_mask}
\includegraphics[width=0.26\linewidth,height=0.12\linewidth]{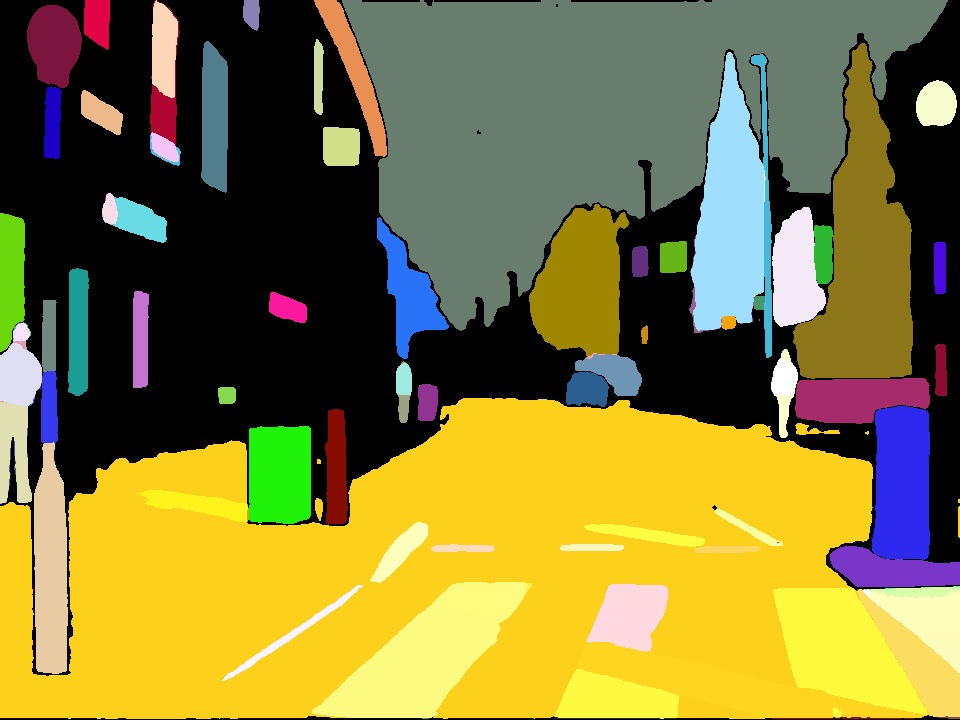}}
\hspace{0.3cm}
\subfloat[LAM's annotation (Ours)]{
\label{Fig.our_anno}
\includegraphics[width=0.26\linewidth,height=0.12\linewidth]{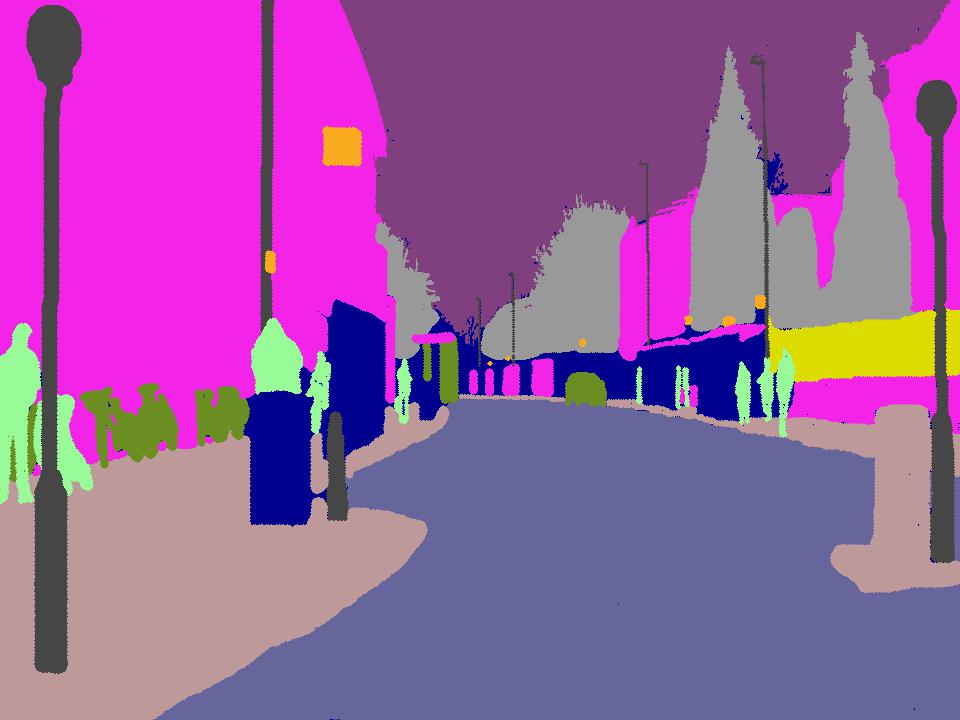}}
\vspace{-0.2cm}
\caption{Comparison of LAM's annotation against SAM's segmentation. \textbf{(a)} Raw image. \textbf{(b)} Illustration of SAM's weaknesses, such as class-agnostic issue (for example, trees are assigned to different semantic IDs (colors)) and coarser annotation (for instance, tree leaves are not annotated well). \textbf{(c)} LAM's annotation overcomes the weaknesses of SAM's segmentation.}
\label{Fig.SAM_reason_to_cluster}
\vspace{-0.8cm}
\end{figure*}

Although recently emerged Segment Anything Model (SAM) \cite{segment-anything} can segment objects automatically from RGB images to serve as labels, it suffers from class-agnostic issue and coarser segmentation (illustrated in \Cref{Fig.SAM_generated_mask}), which hinders the wide application of SAM-driven data annotation in the context of AD. In addition, SAM involves heavy computational load for segmenting each image. For example, it takes 5.34s to segment the image in \Cref{Fig.row_image}. Worse still, SAM's segmentation results are dependent on manually crafted prompts. In general, higher quality prompts enable better segmentation. This necessity of prompts poses a significant challenge when dealing with diverse or rapidly evolving data.

\begin{figure*}[tp]
\centering
\vspace{-0.3cm}
\includegraphics[width=\linewidth,height=0.6\linewidth]{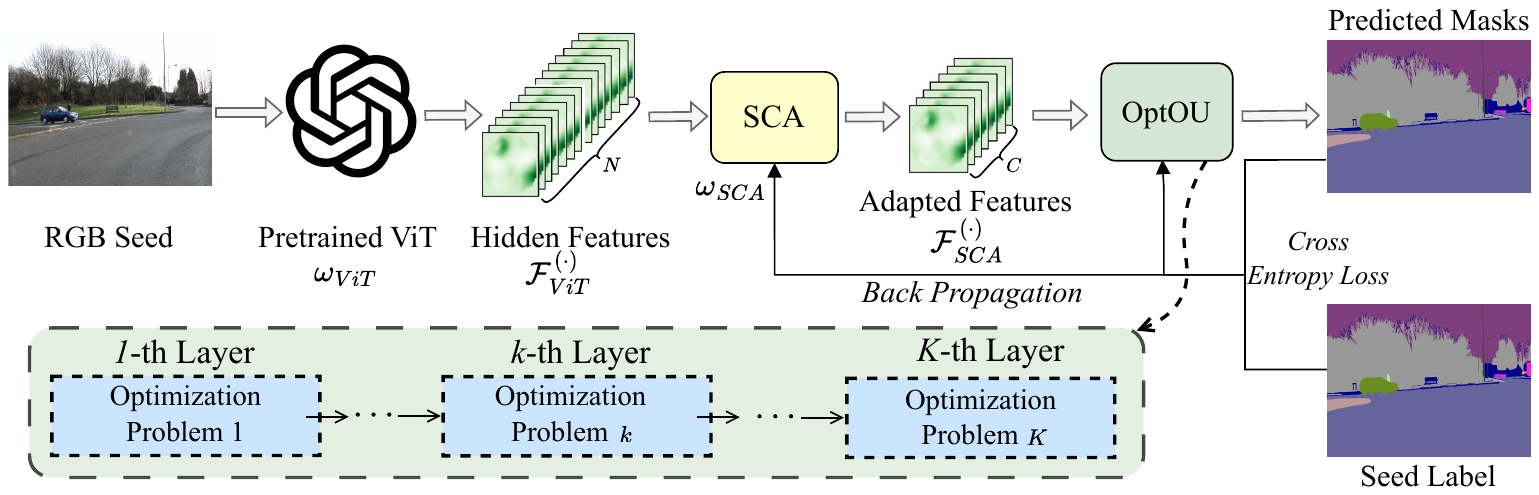}
\vspace{-6.7cm}
\caption{Overview of the proposed LAM.}
\label{fig:overview}
\vspace{-0.7cm}
\end{figure*}

To surmount such SAM's limitations, this paper presents a \underline{L}abel \underline{A}nything \underline{M}odel (denoted as LAM), which is featured as being interpretable, high-fidelity, and prompt-free. LAM consists of following three elements: (I) Pretrained Vision Transformer (ViT) \cite{dosovitskiy2020image}, which acts as a backbone to extract the latent features of each input RGB image. Generally, these features are advanced and learned representations of the input image, and capture both local and global information that is necessary for the subsequent task at hand. (II) Semantic Class Adapter (SCA), which fuses the ViT-extracted hidden features into $C$ channels, where $C$ is the number of semantic classes in the dataset. We implement SCA by using one-layer conv1x1, involving a quite small number of tunable parameters. (III) Optimization-Oriented Unrolling algorithm (OptOU), which aims at optimizing SCA's output so that LAM can generate high-fidelity annotations. Specifically, OptOU contains multiple cascading layers, where within each layer we designate a optimization problem to position the output and the ground truth to be as aligned as possible. Then, we propose to learn OptOU's hyperparameters by converting the multiple cascading optimizations to a learning problem. Since both SCA and OptOU contain a quite small number of trainable parameters, we just provide a single pre-annotated RGB seed image to train them. Once SCA and OptOU are trained, LAM can response diverse and evolving data properly. 
The proposed LAM is outlined in \Cref{fig:overview}.

Benefiting from the proposed LAM, the annotations are class-aware and finer (illustrated in \Cref{Fig.our_anno}). Moreover, LAM enables faster annotation for each RGB image. LAM just takes 0.12s to annotate the image in \Cref{Fig.row_image}, demonstrating its superiority relative to SAM's 5.34s. Most importantly, LAM does not necessarily need image-wise crafted prompts. Instead, it just needs a pre-annotated RGB seed to train SCA and OptOU, which looks feasible and scalable in practice.

Our main contributions are highlighted as follows:
\begin{itemize}
    \item This work proposes an interpretable, high-fidelity, and prompt-free annotator LAM, which incorporates a pretrained ViT along with the proposed SCA and OptOU to generate high-fidelity semantic annotations. 
    \item The proposed OptOU is characterized by an elaborated unrolling optimization mechanism, and can be learned using only one single pre-annotated RGB seed image. 
    \item Extensive experiments on real-world datasets and CARLA \cite{dosovitskiy2017carla} simulation dataset show LAM can generate high-fidelity annotations (almost 100\% in mIoU). 
\end{itemize}

\vspace{-0.1cm}
\section{Related Work}
\vspace{-0.1cm}
\label{related_work}
\subsection{Vision Foundation Models}
Vision-Language Models (VLMs) \cite{ALIGN} are trained using massive datasets that pair images with text. These models utilize distinct encoders for image modality and text modality to generate embeddings, respectively. During training, they employ a contrastive learning objective to enhance the alignment of embeddings from positively correlated image-text pairs. A primary use of these models is in tasks such as zero-shot image-text retrieval or zero-shot classification through textual prompts \cite{CLIP}. Additionally, models like ViLT~\cite{VILT}, VLMo~\cite{VLMO}, and BLIP~\cite{BLIP} have been designed to enhance zero-shot capabilities of visual question answering and image captioning. Methods such as LiT~\cite{LIT}, and BLIP-2~\cite{BLIP2} have been developed to minimize the training expenses for CLIP-like architectures by utilizing pre-trained unimodal models. SAM \cite{segment-anything} can produce segmentation masks effectively, while it is hindered in widely practical use owing to prompt necessity, coarser segmentation, and slow segmentation speed. To tackle SAM's such weaknesses, this paper proposes an interpretable, high-fidelity and prompt-free annotator LAM by leveraging an unrolling optimization mechanism and a single pre-annotated RGB seed image. 

\subsection{Prompt Engineering}
\vspace{-0.1cm}
Prompt engineering \cite{tonmoy2024comprehensive} has become a pivotal strategy for boosting the functionality of pre-trained large language models (LLMs) \cite{10265134} and VLMs \cite{10611726}. It entails the deliberate creation of task-specific directives (known as prompts) to direct the output of models without modifying their parameters. Prompt engineering is particularly prominent in enhancing the flexibility of LLMs and VLMs \cite{10610948}, which allows these models to perform excellently across a variety of tasks and fields. This flexibility marks a departure from conventional methods that typically require retraining or extensive fine-tuning for specific task. As prompt engineering continues to evolve, ongoing research continually uncovers new methods \cite{zhang2024extracting} and applications \cite{xiao2024efficient}. Current researches on prompt engineering include a range of techniques, such as zero-shot prompts \cite{allingham2023simple}, few-shot prompts \cite{lu2021fantastically}, etc. Despite such achievements of prompt engineering, crafting effective prompts often requires deep expertise not only in the model's workings but also in the specific domain knowledge \cite{liu2022design}. This can limit the application of prompt engineering. To surmount such prompt's limitations, this paper proposes to provide a pre-annotated RGB seed image instead of image-specific prompts to enhance the quality of annotations.

\section{Methodology}
\label{methodology}
LAM consists of a pretrained ViT, the proposed SCA and OptOU, and they are introduced in \Cref{method_A}, \Cref{method_B}, and \Cref{method_C}, respectively. \Cref{method_D} demonstrates how LAM learns from the pre-annotated RGB seed and annotates the remaining unlabeled RGB images.

\subsection{Pretrained Vision Transformer (ViT)}
\label{method_A}
Pretrained Vision Transformer (ViT) \cite{dosovitskiy2020image} is a widely used representation extractor to extract latent features of RGB data. Therefore, in this work, we also select ViT as the backbone (with parameters $\pmb{\omega}_{ViT}$) to process the pre-annotated RGB seed image and the remaining unannotated images. 

We denote the pre-annotated RGB seed image as $\mathcal{D}^{(s)}$ and the $i$-th unannotated RGB image as $\mathcal{D}^{(i)}$ for dataset $\mathcal{D}$. Their corresponding latent features $\mathcal{F}^{(s)}_{ViT}$ and $\mathcal{F}_{ViT}^{(i)}$ is extracted in a zero-shot manner as follow: 
\begin{align}
\mathcal{F}^{(\cdot)}_{ViT} = \pmb{\omega}_{ViT}(\mathcal{D}^{(\cdot)}).
\label{Eq:ViT_seed_forward}
\end{align}
$\mathcal{F}^{(\cdot)}_{ViT}$ \footnote{$~*^{(\cdot)}$ can be $*^{(s)}$ or $*^{(i)}$ hereafter, where $*$ represents any related notations.} is a high-dimensional vector, representing the ViT's understanding of the content of $\mathcal{D}^{(\cdot)}$. 

\subsection{Semantic Class Adapter (SCA)}
\label{method_B}
Once $\mathcal{F}_{ViT}^{(\cdot)}$ has been extracted by the ViT, it is transmitted to SCA (with parameters $\pmb{\omega}_{SCA}$) to adapt the ViT's $N$-channel features to $C$ channels (illustrated in \Cref{fig:overview}), where $C$ represents the number of semantic classes in the dataset. SCA's transformation is formulated as follow:  
\begin{align}
\mathcal{F}^{(\cdot)}_{SCA} &= \pmb{\omega}_{SCA}(\mathcal{F}^{(\cdot)}_{ViT}).
\label{Eq:SCA_seed_forward}
\end{align}
$\mathcal{F}_{SCA}^{(\cdot)}$ is then transferred to the downstream OptOU as input for the optimization.

In addition, for the architecture of SCA, we propose to use one-layer conv1x1 coupled with ReLU activation. The rationale behind it is that conv1x1 contains a small number of learnable parameters, which leads to fast convergence and the need of a small size of training data. Specifically, the number of tunable parameters in SCA can be formulated as
\begin{align}
N_{SCA} = (N \times 1 \times 1 + 1) \times C = (N + 1) \times C.
\label{Eq:conv_param}
\end{align}
where $N$ is the number of channels of $\mathcal{F}_{ViT}^{(\cdot)}$; $C$ is the number of channels of $\mathcal{F}_{SCA}^{(\cdot)}$; the "$1 \times 1$'' is the product of the width and height of the kernel; the "+1" accounts for the bias term. 

\subsection{Optimization-Oriented Unrolling Algorithm (OptOU)}
\label{method_C}
OptOU is proposed to generate high-fidelity annotations and is optimized using $\mathcal{D}^{(s)}$ and its ground truth $\mathcal{G}^{(s)}$. 

\begin{figure*}[tp]
\vspace{-0.3cm}
\hspace{-0.9cm}
\includegraphics[width=1.09\linewidth, height=0.75\linewidth]{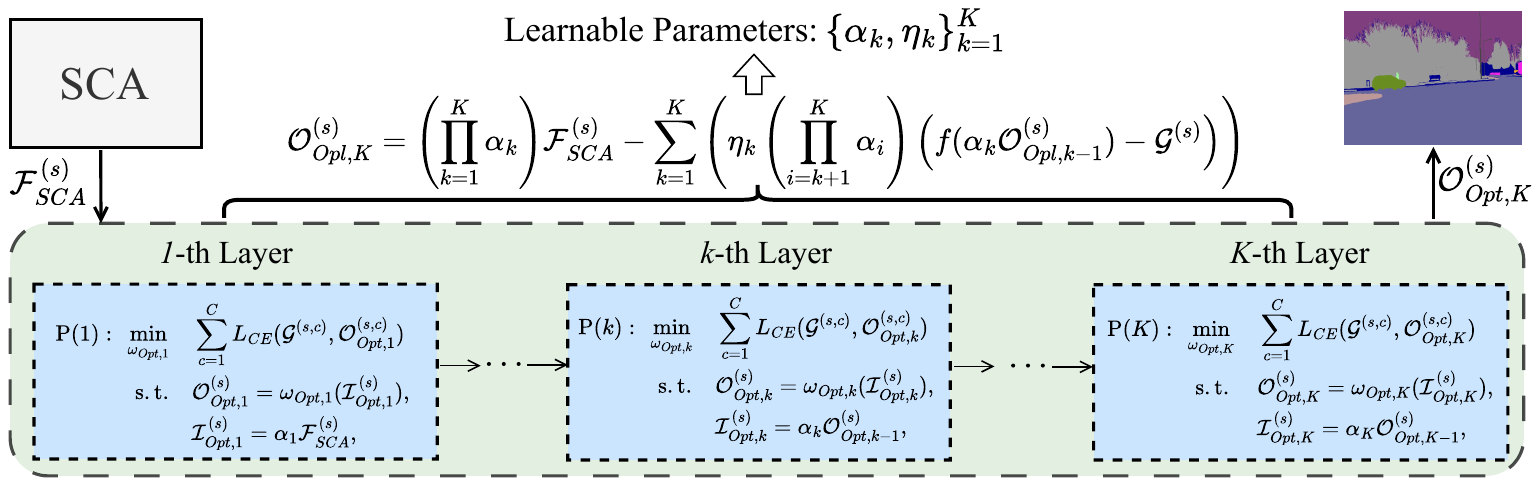}
\vspace{-8.1cm}
\caption{Illustration of the proposed OptOU.}
\label{fig:overview_of_OptOU}
\vspace{-0.6cm}
\end{figure*}

OptOU contains $K$ cascading layers. For the $k$-th layer (where $k \in \{1, 2, \cdots, K\}$), we firstly define the mapping $\pmb{\omega}_{Opt, k}$ of the input $\mathcal{I}_{Opt, k}^{(s)}$ to the ouput $\mathcal{O}_{Opt, k}^{(s)}$, \ie, $\mathcal{O}_{Opt, k}^{(s)} = \pmb{\omega}_{Opt, k}(\mathcal{I}_{Opt, k}^{(s)})$ \footnote{~For the first layer, its input comes from the output of SCA, \ie, $\mathcal{I}_{Opt, 1}^{(s)} = \alpha_1 \mathcal{F}_{SCA}^{(s)}$}. $\mathcal{G}^{(s)}$, $\mathcal{I}_{Opt, k}^{(s)}$ and $\mathcal{O}_{Opt, k}^{(s)}$ all contain $C$ channels, where $c$ indicates the channel order.
Then, we propose to minimize the distance between the RGB seed's ground truth $\mathcal{G}^{(s)}$ and the output $\mathcal{O}_{Opt, k}^{(s)}$ as follows: 
\begin{subequations}
\begin{alignat}{2}
\text{P}(k): ~~&\min_{\pmb{\omega}_{Opt, k}} \quad && \sum\nolimits_{c=1}^C L_{CE}(\mathcal{G}^{(s, c)}, \mathcal{O}_{Opt, k}^{(s, c)}) \\
&~~\mathrm{s.t.} \quad && \mathcal{O}_{Opt, k}^{(s)} = \pmb{\omega}_{Opt, k}(\mathcal{I}_{Opt, k}^{(s)}), \label{eq:constraint1} \\
&&& \mathcal{I}_{Opt, k}^{(s)} = \alpha_k \mathcal{O}^{(s)}_{Opt, k-1}, \label{eq:constraint2}
\end{alignat}
\end{subequations}
where $L_{CE}(\cdot, \cdot)$ is the cross entropy loss; $\alpha_k$ is the scaling factor of the input of $k$-th layer relative to the output of $(k\!-\!1)$-th layer. In conclusion, the proposed OptOU is illustrated in \Cref{fig:overview_of_OptOU}.

For such cascading optimization problems \{P(1), P(2), $\cdots$, P($k$), $\cdots$, P($K$)\}, we propose to solve them together instead of solving them one by one. Specifically, we can regard the optimization problem P($k$) as the $k$-th iteration of a global optimization P. This implies that, for the $k$-th layer, the output $\mathcal{O}_{Opt, k}^{(s)}$ is equal to the input $\mathcal{I}_{Opt, k}^{(s)}$ to substract the gradient of $L_{CE}(\cdot, \cdot)$ with respect to the input $\mathcal{I}_{Opt, k}^{(s)}$.
Therefore, we can use the gradient descent update with step size $\eta_k$ to serve as $\pmb{\omega}_{Opt, k}$, \ie,
\begin{equation}
\small
\begin{aligned}
\hspace{-0.25cm}\mathcal{O}_{Opt, k}^{(s)} \!&=\! \pmb{\omega}_{Opt, k}(\mathcal{I}_{Opt, k}^{(s)}) \\ 
&=\! \mathcal{I}_{Opt, k}^{(s)} \!-\! \eta_k \nabla_{\mathcal{I}_{Opt, k}^{(s)}} \vspace{-0.5cm}L_{CE}(\mathcal{G}^{(s)}, \mathcal{I}^{(s)}_{Opt, k})
\\
&= \alpha_k\mathcal{O}^{(s)}_{Opt, k-1} \!-\! \eta_k \nabla_{\alpha_k\mathcal{O}^{(s)}_{Opt, k\!-\!1}} \hspace{-0.5cm}L_{CE}(\mathcal{G}^{(s)}, \alpha_k \mathcal{O}^{(s)}_{Opt, k\!-\!1}), 
\label{Eq:descent_update}
\end{aligned}
\vspace{-0.4cm}
\end{equation}
where
\begin{align}
    &\hspace{-0.2cm}L_{CE}(\mathcal{G}^{(s)}, \alpha_k\mathcal{O}^{(s)}_{Opt, k-1})\! =\! -\!\sum_{c=1}^C \mathcal{G}^{(s, c)} f(\alpha_k\mathcal{O}^{(s, c)}_{Opt, k-1}), 
    \label{Eq:def_lce}
    \\
    &f(\alpha_k\mathcal{O}^{(s, c)}_{Opt, k-1}) = \log \left(\frac{\exp(\alpha_k\mathcal{O}^{(s, c)}_{Opt, k-1})}{\sum_{c=1}^C \exp(\alpha_k\mathcal{O}^{(s, c)}_{Opt, k-1})}\right).
\label{Eq:intermedia_f} 
\end{align}

Based on above formulation of $L_{CE}(\mathcal{G}^{(s)}, \alpha_k\mathcal{O}^{(s)}_{Opt, k-1})$, we can calculate the gradients by applying the chain rule, \ie, $\nabla_{\alpha_k\mathcal{O}^{(s)}_{Opt, k-1}} \hspace{-0.4cm}L_{CE}(\mathcal{G}^{(s)}, \alpha_k\mathcal{O}^{(s)}_{Opt, k-1}) = $
\begin{align}
\hspace{-0.2cm}-\sum_{c=1}^C \frac{\partial L_{CE}(\mathcal{G}^{(s, c)}, \alpha_k\mathcal{O}^{(s, c)}_{Opt, k-1})}{\partial f(\alpha_k\mathcal{O}^{(s, c)}_{Opt, k-1})} \frac{\partial f(\alpha_k\mathcal{O}^{(s, c)}_{Opt, k-1})}{\partial (\alpha_k\mathcal{O}^{(s, c)}_{Opt, k-1})},
\label{Eq:LCE_grad}
\end{align}
where
\begin{align}
\frac{\partial L_{CE}(\mathcal{G}^{(s, c)}, \alpha_k\mathcal{O}^{(s, c)}_{Opt, k-1})}{\partial f(\alpha_k\mathcal{O}^{(s, c)}_{Opt, k-1})}
= \frac{\mathcal{G}^{(s, c)}}{f(\alpha_k\mathcal{O}^{(s, c)}_{Opt, k-1})},
\label{Eq:L_to_PX}
\\
\hspace{-0.30cm}\frac{\partial f(\alpha_k\mathcal{O}^{(s, c)}_{Opt, k\!-\!1})}{\partial (\alpha_k\mathcal{O}^{(s, c)}_{Opt, k\!-\!1})}
\!=\! f(\alpha_k\mathcal{O}^{(s, c)}_{Opt, k\!-\!1}) (1 \!-\! f(\alpha_k\mathcal{O}^{(s, c)}_{Opt, k\!-\!1})).
\label{Eq:PX_to_logits}
\end{align}
Substituting \Cref{Eq:intermedia_f,Eq:L_to_PX,Eq:PX_to_logits} into \Cref{Eq:LCE_grad}, we can conclude that $\nabla_{\alpha_k\mathcal{O}^{(s)}_{Opt, k-1}} \hspace{-0.4cm} L_{CE}(\mathcal{G}^{(s)}, \alpha_k\mathcal{O}^{(s)}_{Opt, k-1}) = $
\begin{align}
\left(\left(\sum\nolimits_{c=1}^C f(\alpha_k\mathcal{O}^{(s, c)}_{Opt, k-1}) \mathcal{G}^{(s, c)}\right) \!-\!\mathcal{G}^{(s, c)}\right).
\end{align}
Since $\sum_{c=1}^C f(\alpha_k\mathcal{O}^{(s, c)}_{Opt, k-1}) \mathcal{G}^{(s, c)} = f(\alpha_k\mathcal{O}^{(s, c)}_{Opt, k-1})$, the gradients can be simplified to
\begin{equation}
\small    
\hspace{-0.3cm}\nabla_{\alpha_k\mathcal{O}^{(s)}_{Opt, k-1}} \hspace{-0.5cm}L_{CE}(\mathcal{G}^{(s)}, \alpha_k\mathcal{O}^{(s)}_{Opt, k\!-\!1})
\!=\! f(\alpha_k\mathcal{O}^{(s)}_{Opt, k-\!1\!}) \!-\! \mathcal{G}^{(s)}.
\label{Eq:grad_of_LCE}
\end{equation}

Given \Cref{Eq:descent_update}, \Cref{Eq:grad_of_LCE}, and OptOU's cascading nature, the output of OptOU \(\mathcal{O}^{(s)}_{Opt, K}\) can be formulated as 
\begin{align}
&\mathcal{O}^{(s)}_{Opt, K} = \left(\prod_{k=1}^K \alpha_k\right) \mathcal{F}_{SCA}^{(s)} ~- 
\nonumber \\
&\sum_{k=1}^K \left(\eta_k \left(\prod_{i=k+1}^K \alpha_i\right)\left(f(\alpha_k\mathcal{O}^{(s)}_{Opt, k-1}) - \mathcal{G}^{(s)}\right)\right),
\label{Eq:output_of_optou}
\end{align}
where the first term \(\left(\prod_{k=1}^K \alpha_k\right) \mathcal{F}_{SCA}^{(s)}\) scales the SCA's output by the product of all scaling factors (\ie, $\prod_{k=1}^K \alpha_k$); the second term represents the cumulative scaled gradients observed at each step, each also scaled by the subsequent \(\alpha_k\) values from the recent step $k$ to \(K\). 

\Cref{Eq:output_of_optou} implies that the OptOU's optimized output \(\mathcal{O}^{(s)}_{Opt, K}\) depends intricately on the gradient descent path, which is influenced by \(\alpha_k\) and \(\eta_k\). In contradiction to the conventional case that treats $\alpha_k$ and $\eta_k$ as hyperparameters, we propose to treat them as learnable parameters, which improves the predictive accuracy by adapting to extract patterns from data, and avoids manual heuristic search with compromising performance. OptOU contains $K$ layers and each layer contains two learnable parameters, thus, OptOU involves $2K$ trainable parameters in total. 

\subsection{Learning from Seed and Labeling the Unlabeled}
\label{method_D}
Recall that SCA and OptOC have $(N + 1) \times C$ and $2K$ learnable parameters, respectively. Therefore, the total number of learnable parameters in the proposed LAM is
\begin{align}
    N_{Total} = 2K + (N + 1) \times C,
\end{align}
which is a quite small number of learnable parameters compared to the conventional deep neural networks. Theoretically, it requires a small number of training data. Therefore, we propose to use the single pre-annotated RGB seed to train $\pmb{\omega}_{SCA}$ and $\{\alpha_k, \eta_k\}_{k=1}^{K}$ via back propagation, \ie,
\begin{align}
    \pmb{\omega}_{SCA}, \{\alpha_k, \eta_k\}_{k=1}^{K} \gets \min ~ L_{CE}(\mathcal{G}^{(s)}, \mathcal{O}_{Opt, K}^{(s)}).
    \label{Eq:train_learnables}
\end{align}
Once trained, we denote the entire LAM model as
\begin{align}
    \pmb{\omega} = \{\pmb{\omega}_{ViT}, \pmb{\omega}_{SCA}, \{\alpha_k, \eta_k\}_{k=1}^{K}\}.
\end{align}

Once $\pmb{\omega}$ is available, we can annotate unlabeled RGB image $\mathcal{D}^{(i)}$ by assigning labels
\begin{align}
    \mathcal{G}^{(i)} = \arg\max ~ \pmb{\omega}(\mathcal{D}^{(i)}), ~~ i \in \{1, 2, \cdots, |\mathcal{D}|\},
\end{align}

In conclusion, LAM is outlined in Algorithm \ref{Algo:OptOU}.

\begin{algorithm}[tp]
\caption{Label Anything Model (LAM)}
\label{Algo:OptOU}
\SetAlgoLined
\KwIn{$\pmb{\omega}_{ViT}$, $\mathcal{D}^{(seed)}$, $\mathcal{G}^{(seed)}$, $K$, $\mathcal{D}$, $epochs$}
\KwOut{Ground truth $\mathcal{G}^{(\cdot)}$ of $\mathcal{D}^{(\cdot)}$}
\tcp{Learning from seed}
Initialize $\alpha_{k}, \eta_{k} \gets \alpha_0, \eta_0, \ k \in \{1, 2, \cdots K\}$ \\
\For{$epoch\ e \gets 1$ \KwTo $epochs$}{ 
$\mathcal{F}^{(s)}_{ViT} = \pmb{\omega}_{ViT}(\mathcal{D}^{(s)})
\label{Eq:ViT_seed_forward}$ \\
$\mathcal{F}^{(s)}_{SCA} = \pmb{\omega}_{SCA}(\mathcal{F}^{(s)}_{ViT})$ \\
$\mathcal{O}^{(s)}_{Opt, K} \gets$ \Cref{Eq:output_of_optou} \\
$\pmb{\omega}_{SCA}, \{\alpha_k, \eta_k\}_{k=1}^{K} \gets$ \Cref{Eq:train_learnables} \\
}
\tcp{Labeling the unlabeled}
$\pmb{\omega} = \{\pmb{\omega}_{ViT}, \pmb{\omega}_{SCA}, \{\alpha_k, \eta_k\}_{k=1}^{K}\}$ \\ 
\For {$\mathcal{D}^{(i)}  \in \mathcal{D}$}{
$\mathcal{G}^{(i)} = \arg\max ~ \pmb{\omega}(\mathcal{D}^{(i)})$
}
\end{algorithm}

\section{Experiments}
\label{experiments}
In this section, we carry out comprehensive experiments to verify the proposed LAM's performance of labelling unlabeled images. 

\subsection{Datasets, Evaluation Metrics and Implementation}
\subsubsection{Datasets}
The \textbf{Cityscapes} dataset \cite{Cordts2016Cityscapes} consists of 3,475 images with 20 pixel-level classes. The \textbf{CamVid} dataset \cite{brostow2008segmentation} totally includes 701 samples with 12 pixel-level classes. The \textbf{Apolloscape} dataset \cite{wang2019apolloscape} has 1,255 images with 23 pixel-level classes. The \textbf{CARLA\_ADV} dataset captured from CARLA simulator \cite{dosovitskiy2017carla} includes 2,937 images with 23 pixel-level classes, which is specifically designed to encompass a range of adverse weather conditions, such as fog, clouds, rain, darkness, and combinations thereof. These four datasets all cover common traffic objects, such as vehicle, building, tree, pedestrians, etc. Note that images in these four datasets are paired with ground truth. We randomly select one image as the RGB seed to optimize SCA and OptOU. 

\subsubsection{Evaluation Metrics}
We use two commonly employed metrics: mean Intersection over Union (\textbf{mIoU}), which quantifies the overlap between the predicted mask and the ground truth; and mean F1 (\textbf{mF1}), which offers a harmonized measure of both precision and recall.

\subsubsection{Implementation Details}
SAC and OptOU are developed using the Pytorch framework and are trained on two NVIDIA GeForce 4090 GPUs. For optimization, the Adam optimizer is chosen with Betas values of 0.9 and 0.999, and a weight decay of 1e-4. SAC and OptOU are trained with a learning rate of 3e-4. Additionally, the parameter $K$ in OptOU is configured to be 10. In addition, we also compare LAM with other models, such as BiSecNetV2 \cite{yu2021bisenet}, SegNet \cite{badrinarayanan2017segnet}, DeepLabv3+ \cite{chen2018encoderdecoder}, and Segformer \cite{xie2021segformer}, and ViT+ASSP (where ViT serves as backbone while ASSP \cite{chen2017deeplab} serves as downstream perception head).

\subsection{Main Results and Empirical Analysis}
In this section, we will present experimental results on the LAM's labeling performance and conduct additional empirical analyses, from quantitative and qualitative perspectives:

\begin{table*}[tp]
\setlength{\tabcolsep}{2.18pt}
\renewcommand{\arraystretch}{0.8} 
\caption{Quantitative results of LAM's annotation, across multiple datasets, relative to other SOTA competitors}
\begin{tabularx}{\linewidth}{c|ccccc|ccccc|ccccc|ccccc}
\hline
\multirow{2}{*}{\vspace{-0.3cm}Method}                                                  & \multicolumn{5}{c|}{CamVid}                                                                                                                                                                                                                           & \multicolumn{5}{c|}{Cityscapes}                                                                                                                                                                                                                      & \multicolumn{5}{c|}{Apolloscapes}                                                                                                                                                                                                                     & \multicolumn{5}{c}{CARLA\_ADV}                                                                                                                                                                                                                        \\ \cline{2-21} 
                                                                         & \begin{tabular}[c]{@{}c@{}}\# of \\ train\end{tabular} & \begin{tabular}[c]{@{}c@{}}\# of\\  test\end{tabular} & epochs               & \begin{tabular}[c]{@{}c@{}}mIoU\\  (\%)\end{tabular}  & \begin{tabular}[c]{@{}c@{}}mF1\\  (\%)\end{tabular}   & \begin{tabular}[c]{@{}c@{}}\# of \\ train\end{tabular} & \begin{tabular}[c]{@{}c@{}}\# of\\  test\end{tabular} & epochs               & \begin{tabular}[c]{@{}c@{}}mIoU\\  (\%)\end{tabular}  & \begin{tabular}[c]{@{}c@{}}mF1\\  (\%)\end{tabular}  & \begin{tabular}[c]{@{}c@{}}\# of \\ train\end{tabular} & \begin{tabular}[c]{@{}c@{}}\# of\\  test\end{tabular} & epochs               & \begin{tabular}[c]{@{}c@{}}mIoU\\  (\%)\end{tabular}   & \begin{tabular}[c]{@{}c@{}}mF1\\  (\%)\end{tabular}  & \begin{tabular}[c]{@{}c@{}}\# of \\ train\end{tabular} & \begin{tabular}[c]{@{}c@{}}\# of\\  test\end{tabular} & epochs               & \begin{tabular}[c]{@{}c@{}}mIoU\\  (\%)\end{tabular}  & \begin{tabular}[c]{@{}c@{}}mF1\\  (\%)\end{tabular}   \\ \hline
BiSecNetV2                                                               & \multirow{5}{*}{600}                                   & \multirow{5}{*}{101}                                  & \multirow{5}{*}{200} & 47.89                                                 & 53.33                                                 & \multirow{5}{*}{2,975}                                 & \multirow{5}{*}{500}                                  & \multirow{5}{*}{200} & 33.63                                                 & 43.32                                                & \multirow{5}{*}{854}                                   & \multirow{5}{*}{400}                                  & \multirow{5}{*}{200} & 22.92                                                  & 27.12                                                & \multirow{5}{*}{2,763}                                 & \multirow{5}{*}{1,921}                                & \multirow{5}{*}{200} & 28.80                                                 & 33.59                                                 \\
SegNet                                                                   &                                                        &                                                       &                      & 46.60                                                 & 50.18                                                 &                                                        &                                                       &                      & 43.13                                                 & 52.87                                                &                                                        &                                                       &                      & 21.01                                                  & 24.60                                                &                                                        &                                                       &                      & 31.67                                                 & 37.15                                                 \\
DeepLabv3+                                                               &                                                        &                                                       &                      & 69.46                                                 & 77.58                                                 &                                                        &                                                       &                      & 69.04                                                 & 75.95                                                &                                                        &                                                       &                      & 26.58                                                  & 32.32                                                &                                                        &                                                       &                      & 36.90                                                 & 43.78                                                 \\
Segformer                                                                &                                                        &                                                       &                      & 34.23                                                 & 38.86                                                 &                                                        &                                                       &                      & 39.37                                                 & 46.23                                                &                                                        &                                                       &                      & -                                                      & -                                                    &                                                        &                                                       &                      & -                                                     & -                                                     \\
ViT+ASSP                                                                &                                                        &                                                       &                      & 68.12                                                 & 77.01                                                 &                                                        &                                                       &                      & 26.31                                                 & 30.37                                                &                                                        &                                                       &                      & 17.11                                                      & 21.18                                                    &                                                        &                                                       &                      & 31.64                                                     & 37.58                                                     \\ \hline
\textbf{LAM (Ours)} & 1                                                      & 700                                                   & 1                    & \begin{tabular}[c]{@{}c@{}}\textbf{99.99}\end{tabular} & \begin{tabular}[c]{@{}c@{}}\textbf{99.99}\end{tabular} & 1                                                      & 3,474                                                 & 1                    & \begin{tabular}[c]{@{}c@{}}\textbf{99.99}\end{tabular} & \begin{tabular}[c]{@{}c@{}}\textbf{99.99}\end{tabular} & 1                                                      & 1,253                                                 & 1                    & \begin{tabular}[c]{@{}c@{}}\textbf{99.99}\end{tabular} & \begin{tabular}[c]{@{}c@{}}\textbf{99.99}\end{tabular} & 1                                                      & 4,683                                                 & 1                    & \begin{tabular}[c]{@{}c@{}}\textbf{99.99}\end{tabular} & \begin{tabular}[c]{@{}c@{}}\textbf{99.99}\end{tabular} \\ \hline
\end{tabularx}
\label{Tab:quan_perf_LAM}
\vspace{-0.7cm}
\end{table*}

\begin{figure}[tp]
\vspace{-0.1cm}
\centering
\subfloat[mIoU]{\includegraphics[width=0.5\linewidth, height=0.3\linewidth]{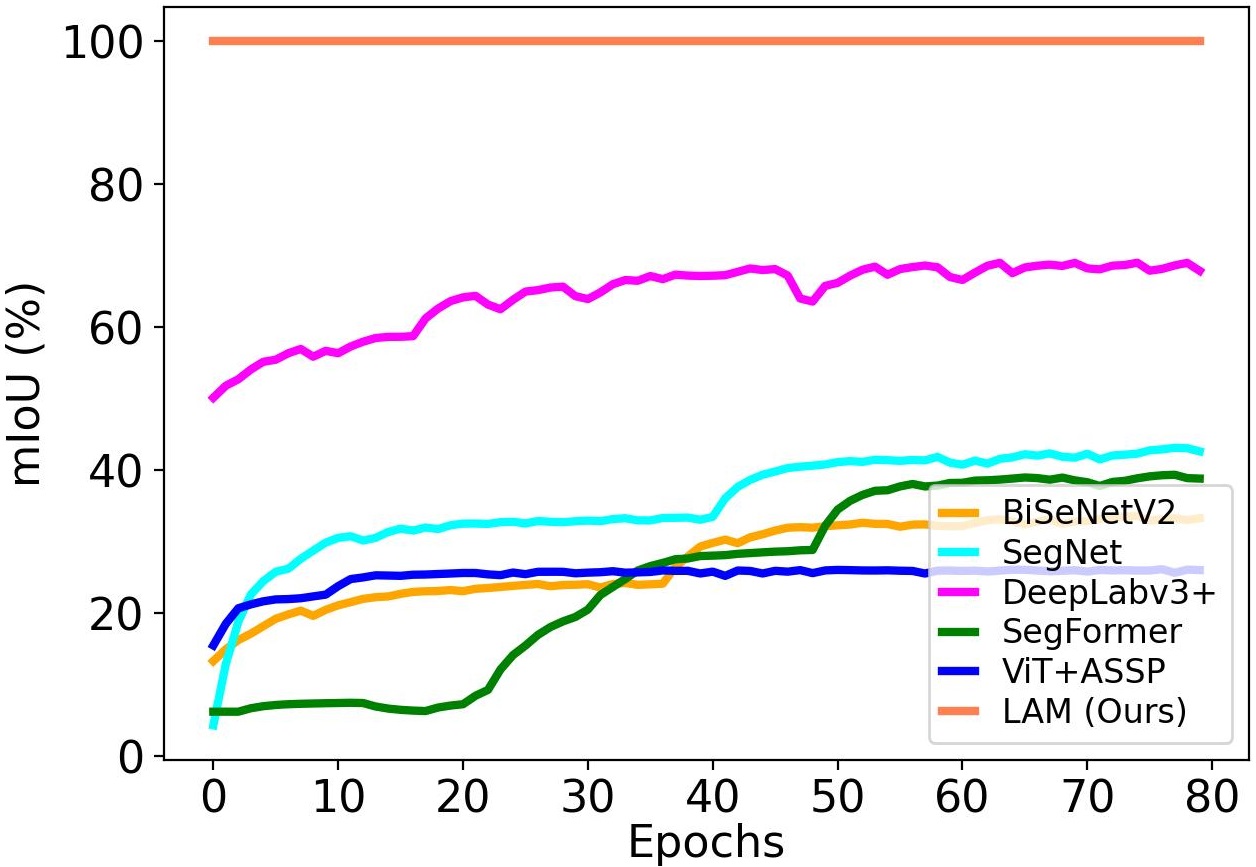}%
\label{Fig:city_mF1}}
\subfloat[mF1]{\includegraphics[width=0.5\linewidth, height=0.3\linewidth]{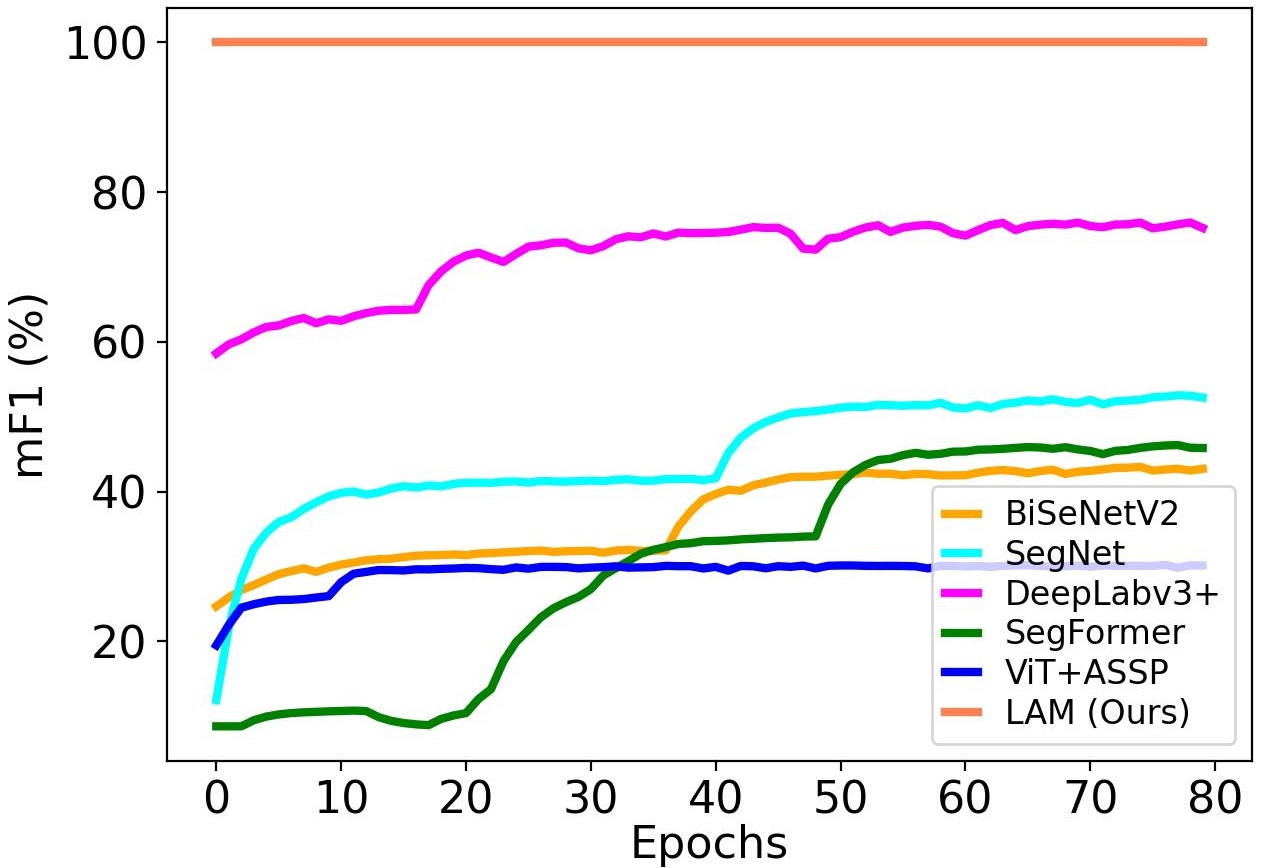}%
\label{Fig:city_mIoU}}
\vspace{-0.2cm}
\caption{Convergence comparison of considered methods.}
\label{Fig:convergece_comparison}
\vspace{-0.4cm}
\end{figure}

\subsubsection{Quantitative Evaluation}

\begin{table}[tp]
\setlength{\tabcolsep}{8.3pt}
\renewcommand{\arraystretch}{0.8} 
\centering
\caption{The comparison of the number of learnable parameters, across multiple datasets}
\begin{tabularx}{\linewidth}{ccccc}
\hline
\multirow{2}{*}{} & \multicolumn{4}{c}{SCA + OptOU}            \\ \cline{2-5} 
                  & $N$   & $C$  & $K$  & \# of Learnable Parameters \\ \hline
CamVid            & 256 & 12 & 10 & 3,104                      \\
Cityscapes        & 256 & 19 & 10 & 4,903                      \\
Apolloscapes      & 256 & 23 & 10 & 5,931                      \\
CARLA\_ADV        & 256 & 23 & 10 & 5,931                      \\ \hline
\end{tabularx}
\label{Tab:learnable_parameters}
\vspace{-0.2cm}
\end{table}

\Cref{Tab:quan_perf_LAM} presents quantitative results of the proposed LAM's annotation performance against other competitors. The evaluation uses above discussed two metrics across CamVid, Cityscapes, Apolloscapes, CARLA\_ADV datasets. We can observe following patterns: (I) LAM achieves almost 100\% on all metrics across all datasets, demonstrating a great superiority over other models. This suggests that LAM consistently produces annotations that perfectly match the ground truth. The reason behind it can be attributed to the proposed OptOU optimization scheme. (II) LAM requires only one pre-annotated RGB seed image to train it, which can be attributed to the quite small number of learnable parameters. (III) LAM converges within one epoch (illustrated in \Cref{Fig:convergece_comparison}), demonstrating faster convergence than other models. This fast convergence property also can be attributed to the quite small volume of trainable parameters.

\Cref{Tab:learnable_parameters} compares the number of learnable parameters for ``SCA + OptOU'' across datasets. It highlights how $N$ (Number of ViT's output channels), $C$ (Dataset's number of semantic classes), and $K$ (Number of OptOU layers) affect the total number of learnable parameters. We can observe following patterns: (I) The number of learnable parameters is primarily influenced by $C$. Datasets with more semantic classes (like Apolloscapes with 23 classes) require more parameters compared to datasets with fewer classes (like CamVid with 12 classes). (II) The number of learnable parameters remains remarkably low, regardless of the dataset used. This small volume of parameters reduces the over-reliance on extensive labeled data. Consequently, even training LAM with a single RGB seed image can achieve excellent performance, highlighting its data efficiency. 

\begin{table*}[tp]
\centering
\renewcommand{\arraystretch}{0.24}
\addtolength{\tabcolsep}{-0.45pt}
\caption{Qualitative performance of LAM across multiple datasets}
\vspace{-0.1cm}
\begin{tabularx}{\linewidth}{|c|c|ccccc@{\hspace{0.13em}}|}
\hline
\multirow{2}{*}{\rotatebox[origin=c]{90}{\parbox[c]{0.6cm}{\textbf{CamVid}}}} &\verticaltext[26pt]{\small}{\textbf{Raw Images}} &\hspace{-0.28cm}
\includegraphics[width=0.182\linewidth, height=0.12\linewidth]{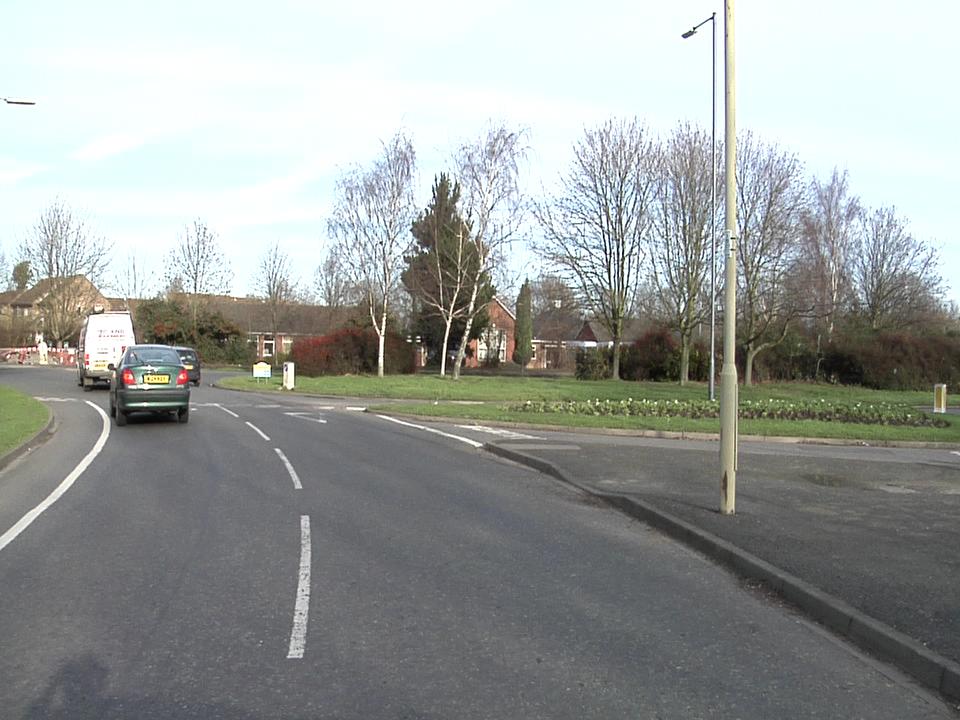} &\hspace{-0.47cm}
\includegraphics[width=0.182\linewidth, height=0.12\linewidth]{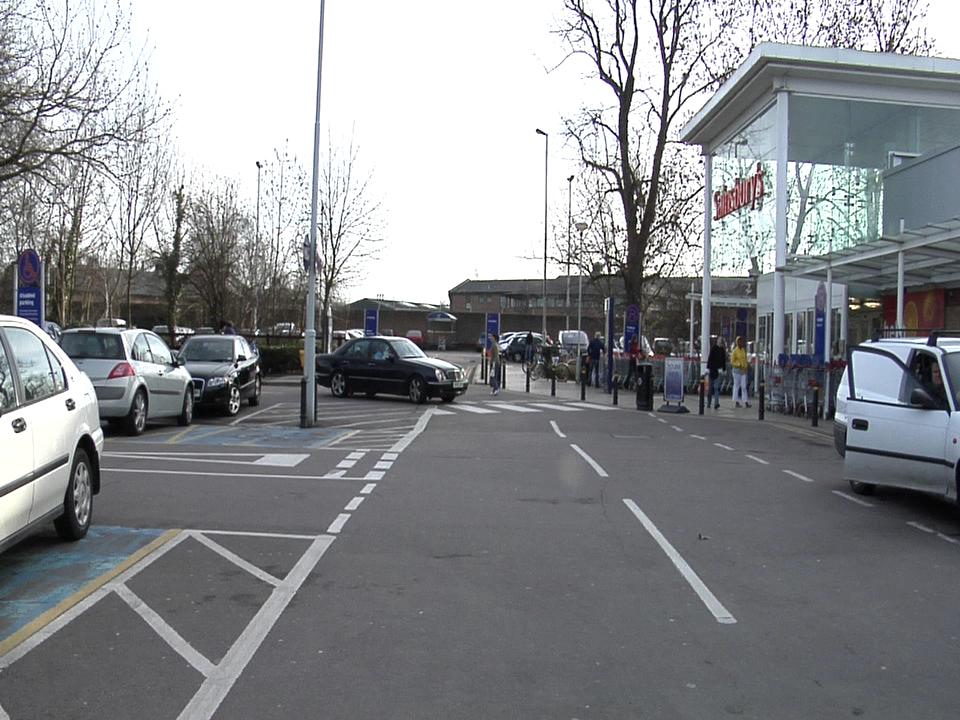} &\hspace{-0.47cm}
\includegraphics[width=0.182\linewidth, height=0.12\linewidth]{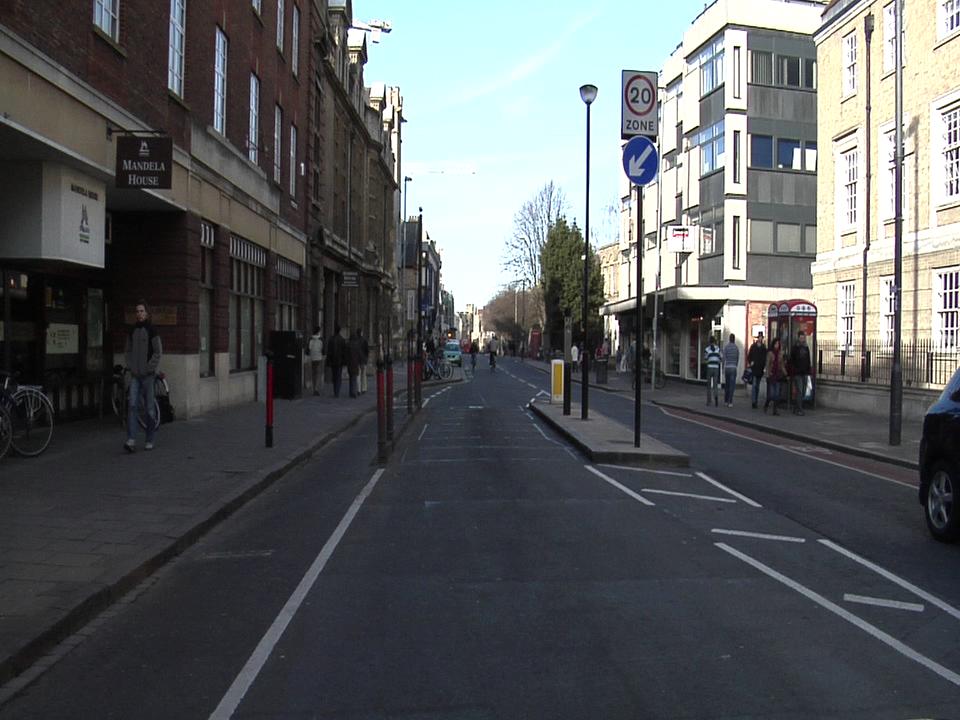} &\hspace{-0.47cm}
\includegraphics[width=0.182\linewidth, height=0.12\linewidth]{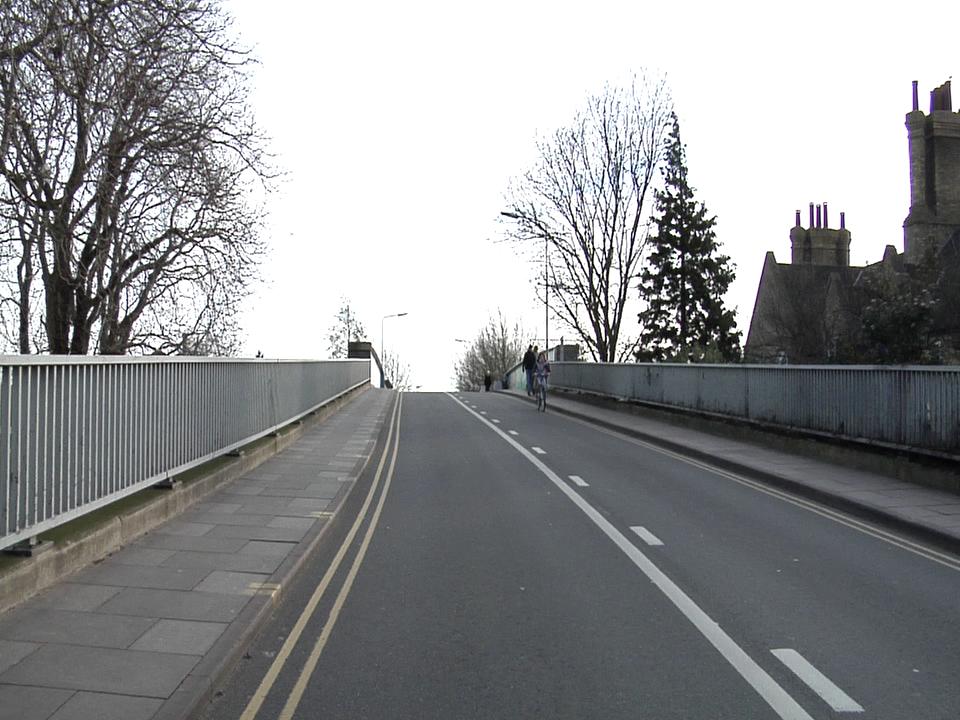} &\hspace{-0.47cm}
\includegraphics[width=0.188\linewidth, height=0.12\linewidth]{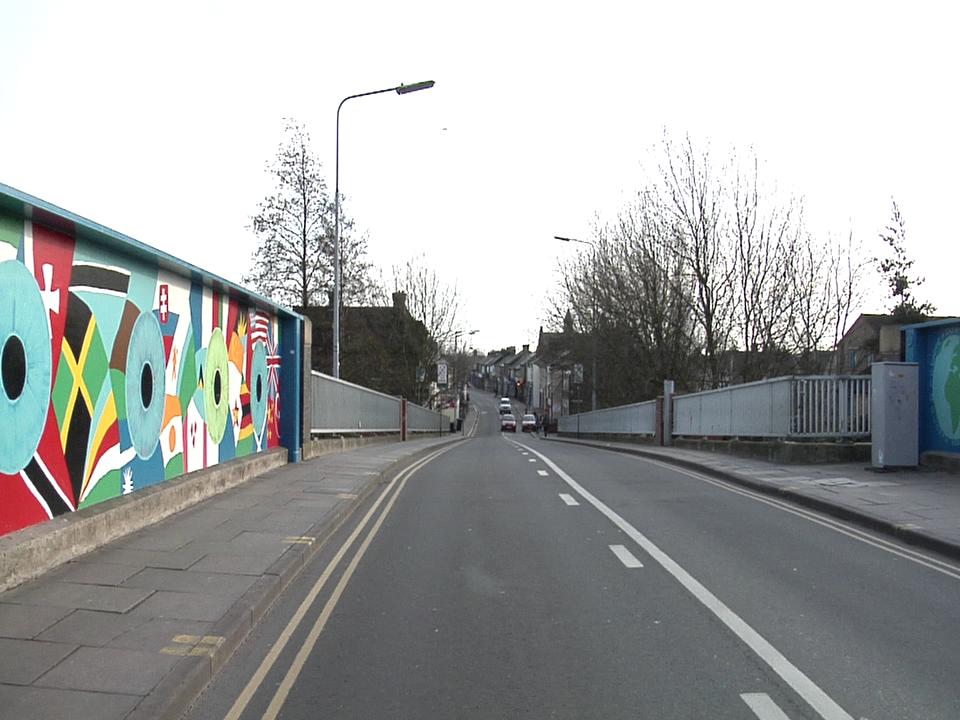}  \\ \cline{2-2}
                  &\verticaltext[26pt]{\small}{\textbf{Annotations}} &\hspace{-0.28cm}
\includegraphics[width=0.182\linewidth, height=0.12\linewidth]{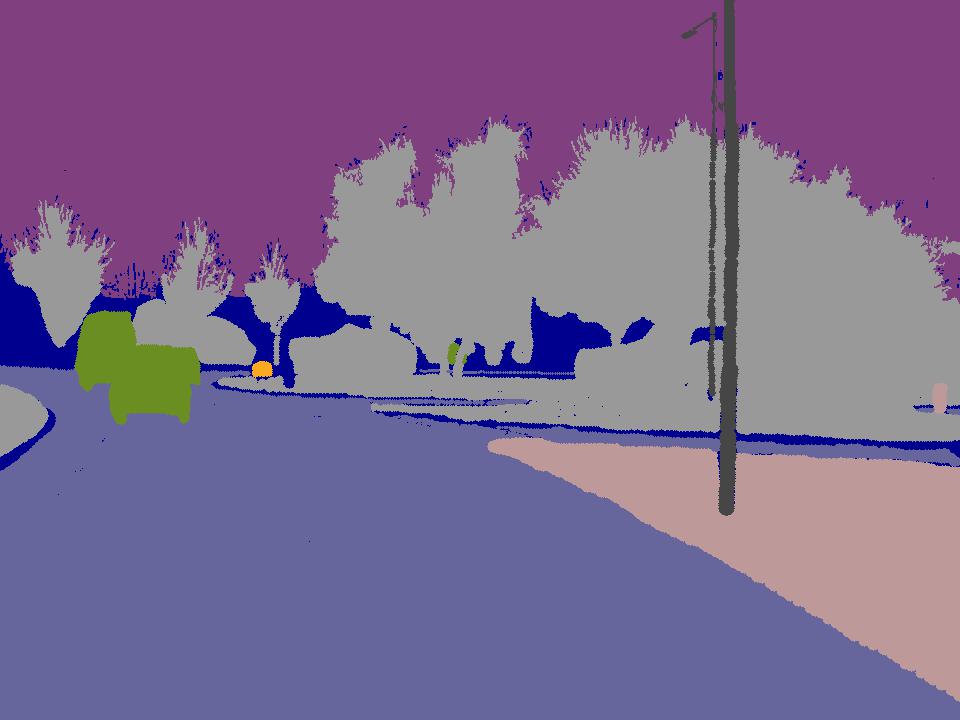} &\hspace{-0.47cm}
\includegraphics[width=0.182\linewidth, height=0.12\linewidth]{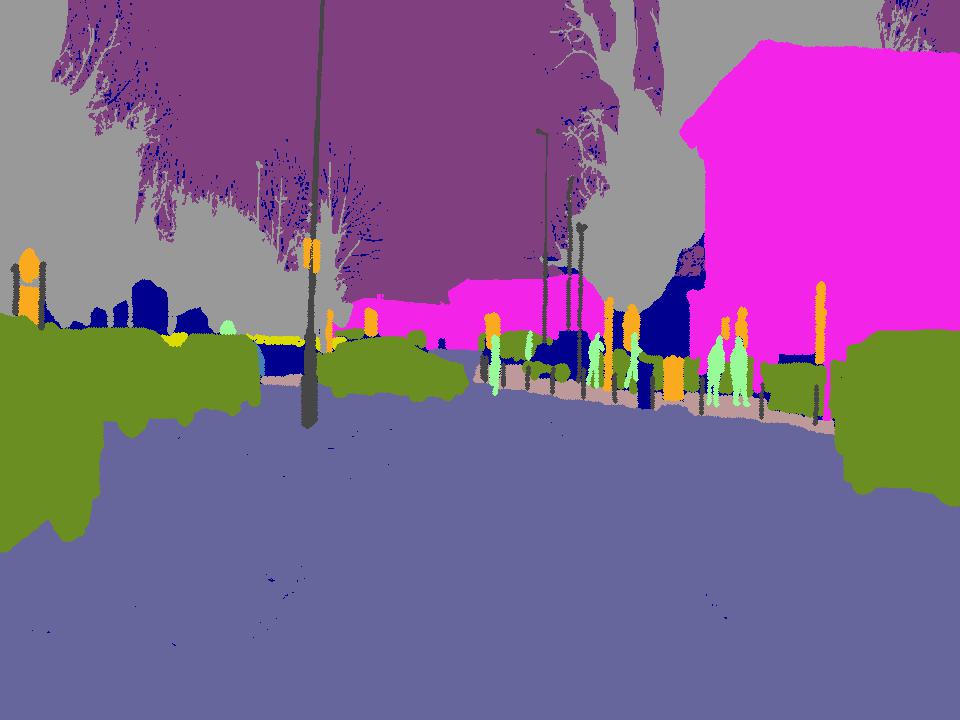} &\hspace{-0.47cm}
\includegraphics[width=0.182\linewidth, height=0.12\linewidth]{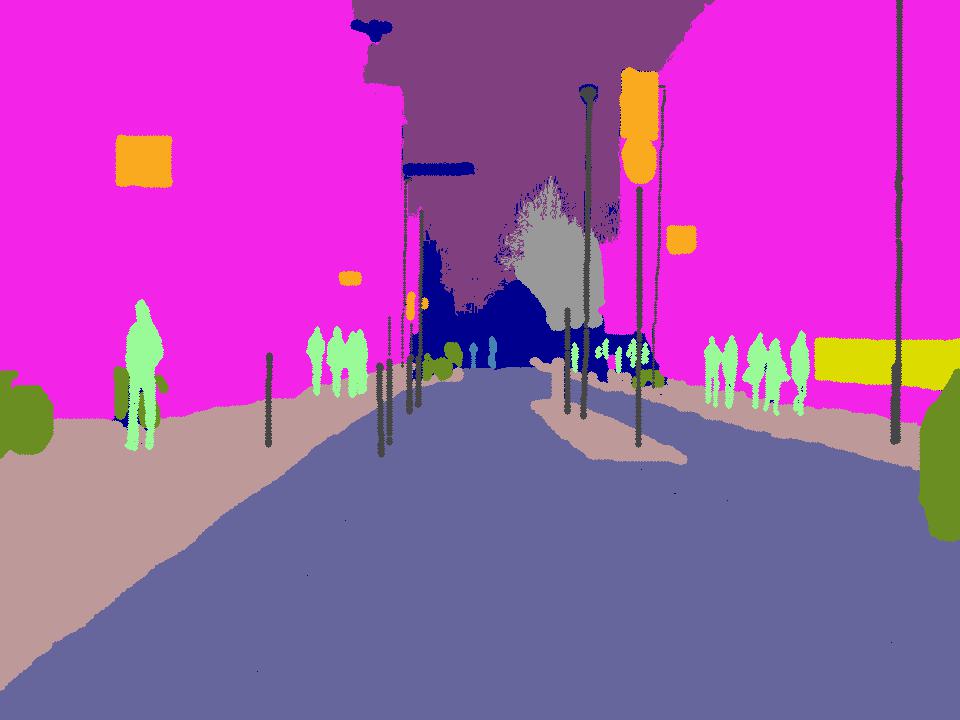} &\hspace{-0.47cm}
\includegraphics[width=0.182\linewidth, height=0.12\linewidth]{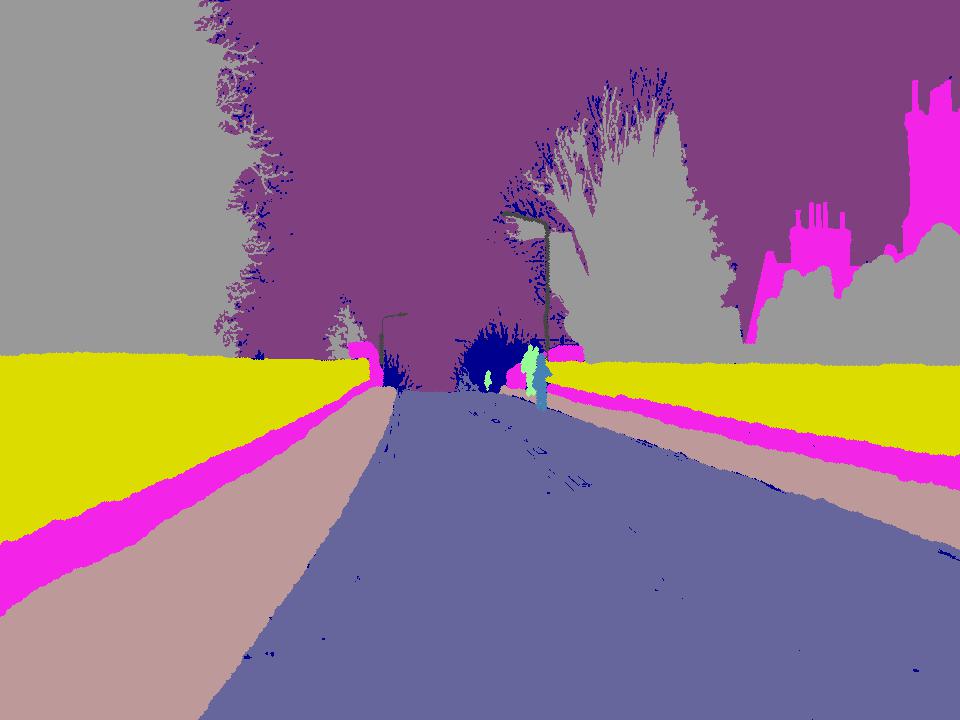} &\hspace{-0.47cm}
\includegraphics[width=0.188\linewidth, height=0.12\linewidth]{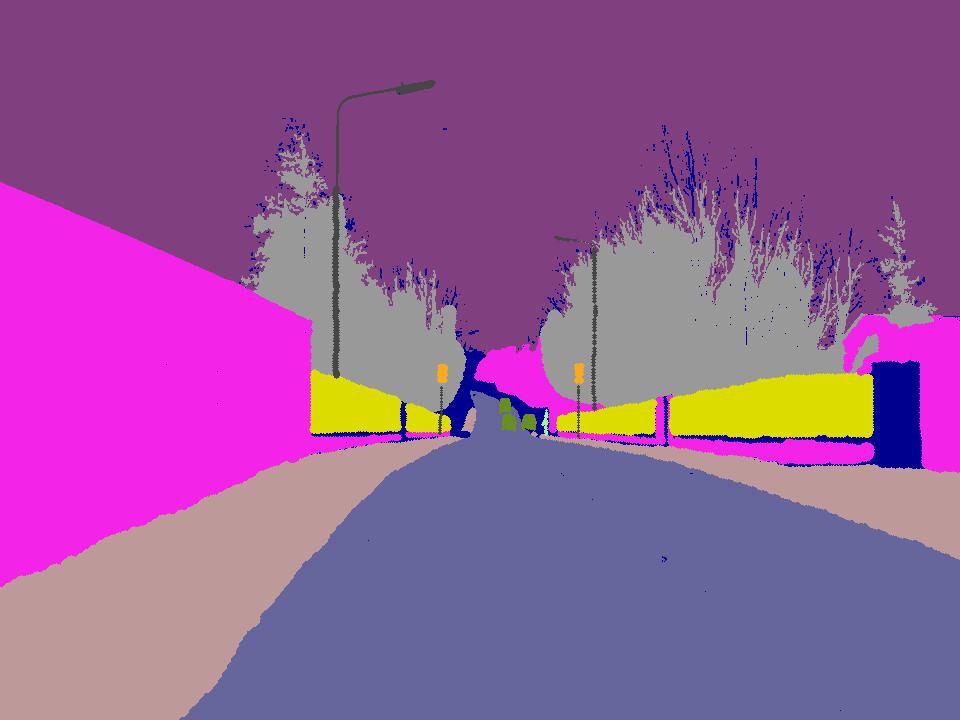}  \\ \cline{1-2}
\multirow{2}{*}{\rotatebox[origin=c]{90}{\parbox[c]{0.7cm}{\textbf{Cityscapes}}}} &\verticaltext[26pt]{\small}{\textbf{Raw Images}} &\hspace{-0.28cm}
\includegraphics[width=0.182\linewidth, height=0.12\linewidth]{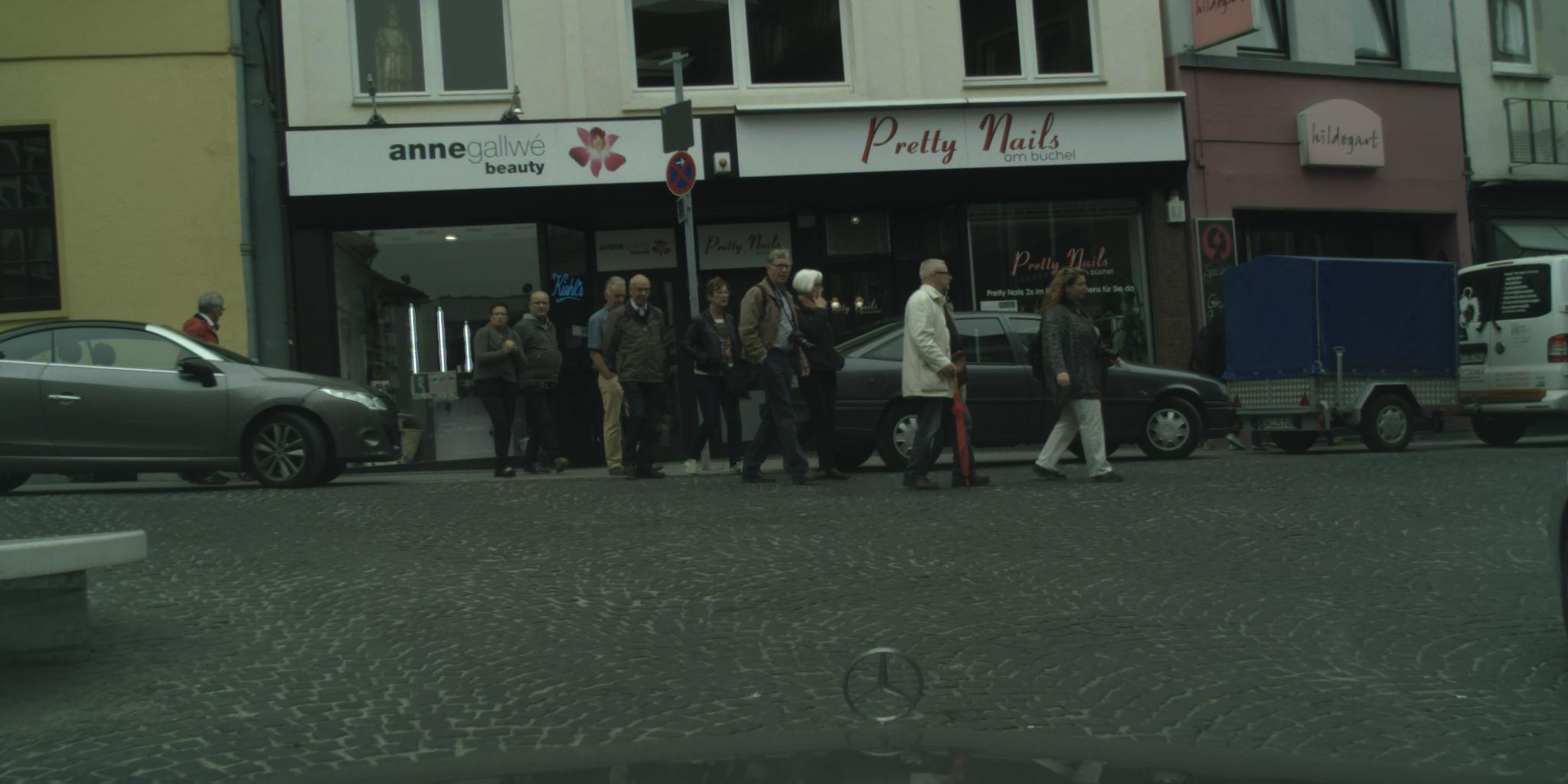} &\hspace{-0.47cm}
\includegraphics[width=0.182\linewidth, height=0.12\linewidth]{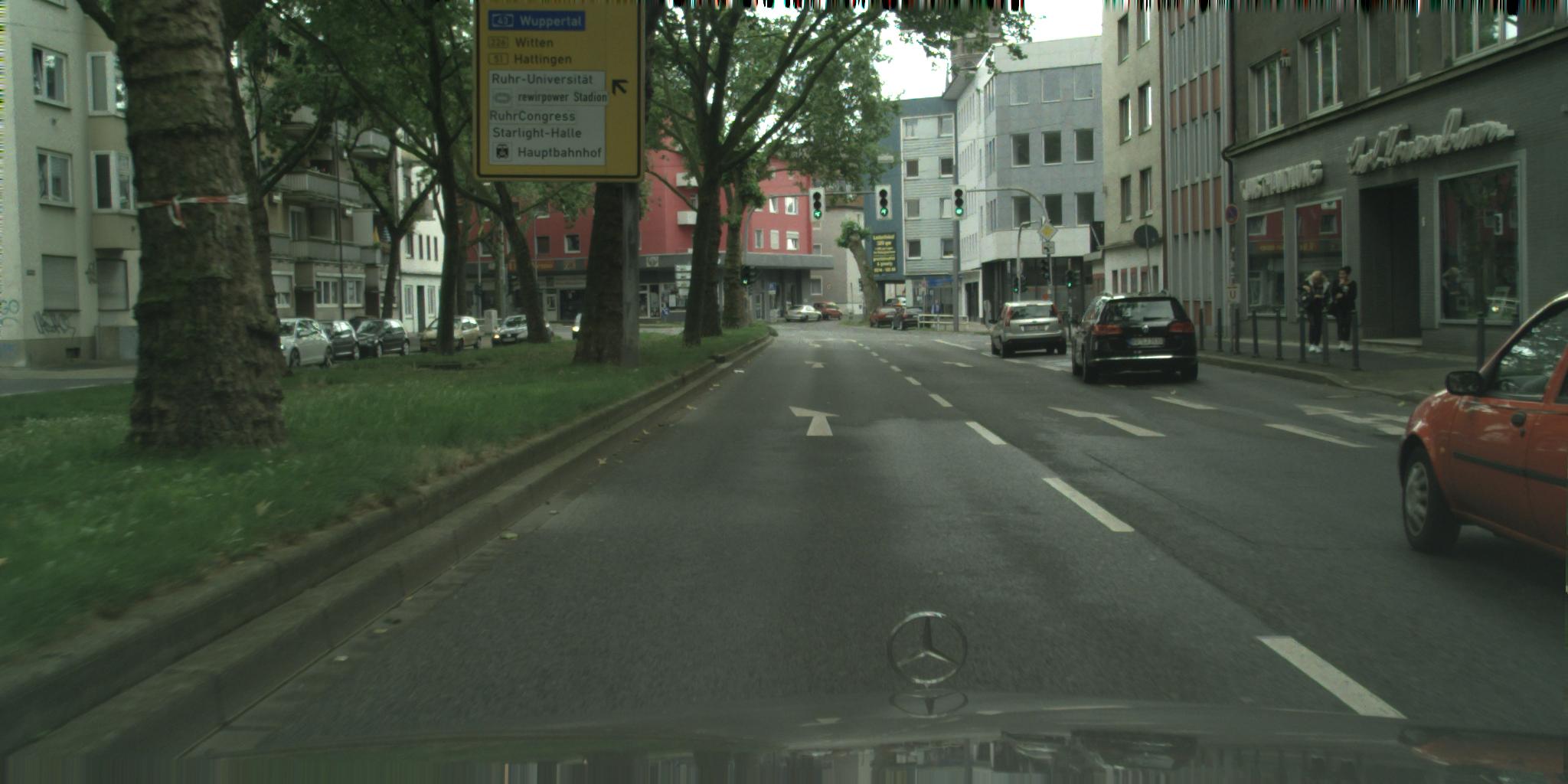} &\hspace{-0.47cm}
\includegraphics[width=0.182\linewidth, height=0.12\linewidth]{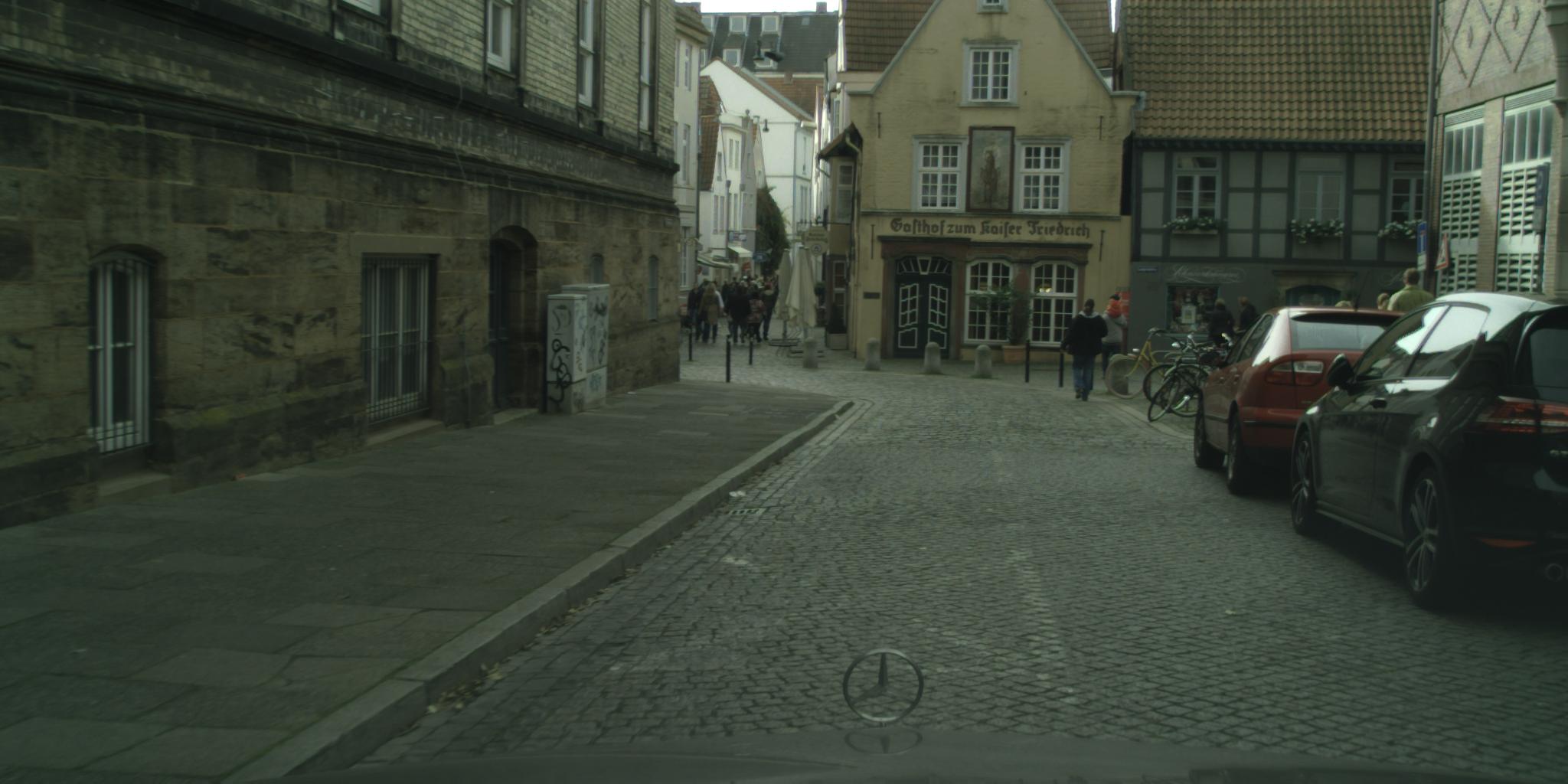} &\hspace{-0.47cm}
\includegraphics[width=0.182\linewidth, height=0.12\linewidth]{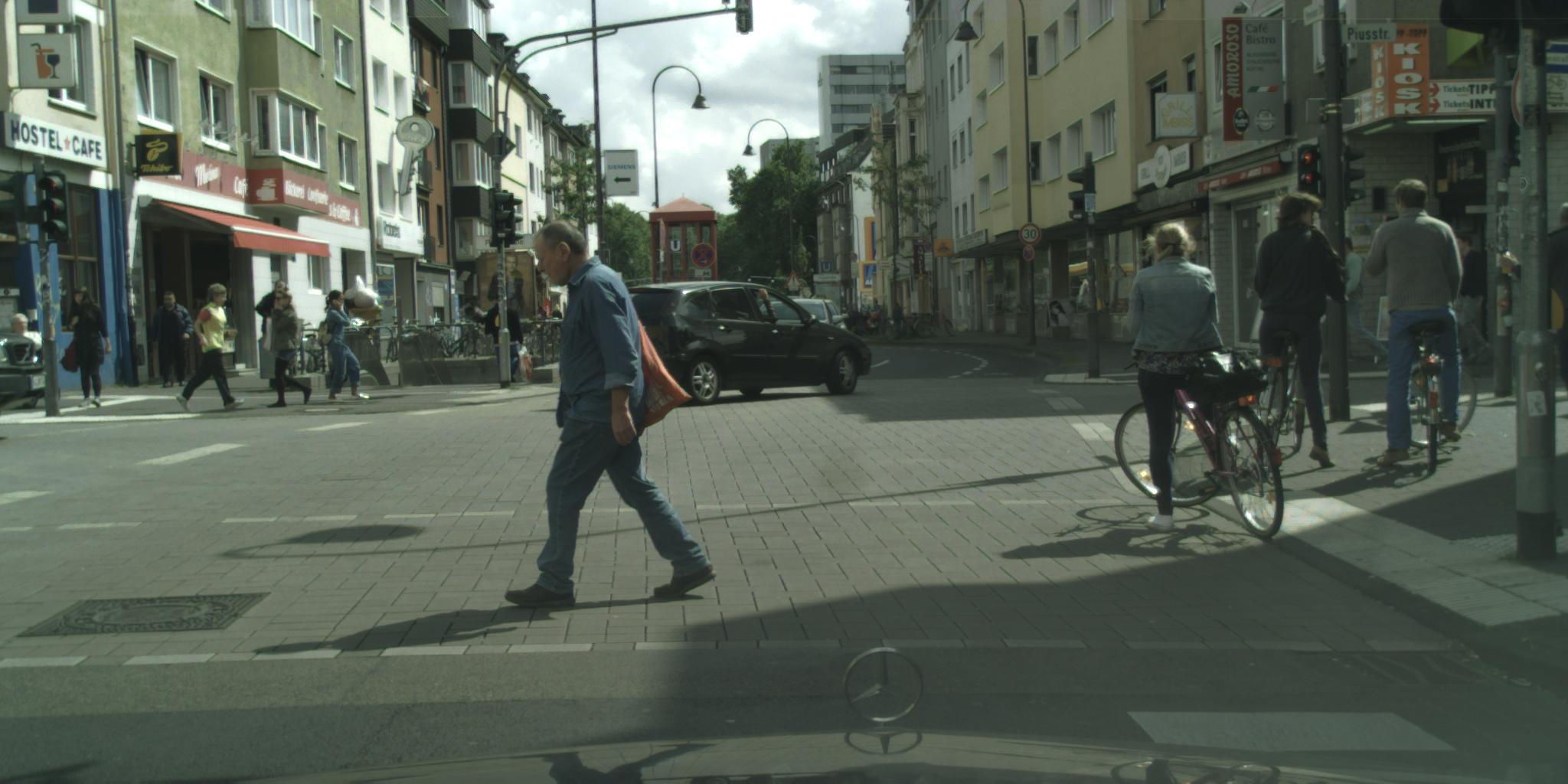} &\hspace{-0.47cm}
\includegraphics[width=0.188\linewidth, height=0.12\linewidth]{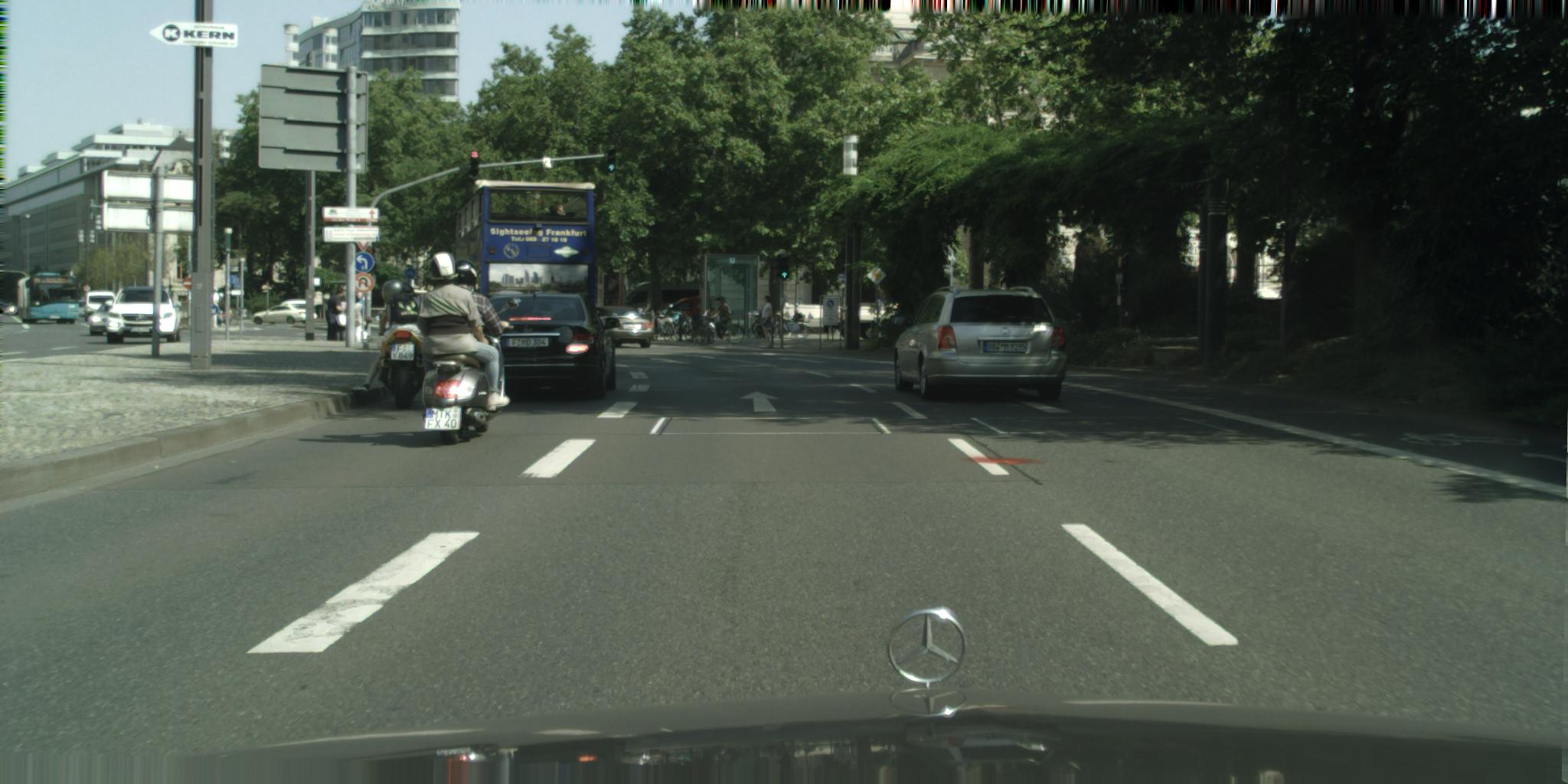} \\ \cline{2-2}
                  &\verticaltext[26pt]{\small}{\textbf{Annotations}} &\hspace{-0.28cm}
\includegraphics[width=0.182\linewidth, height=0.12\linewidth]{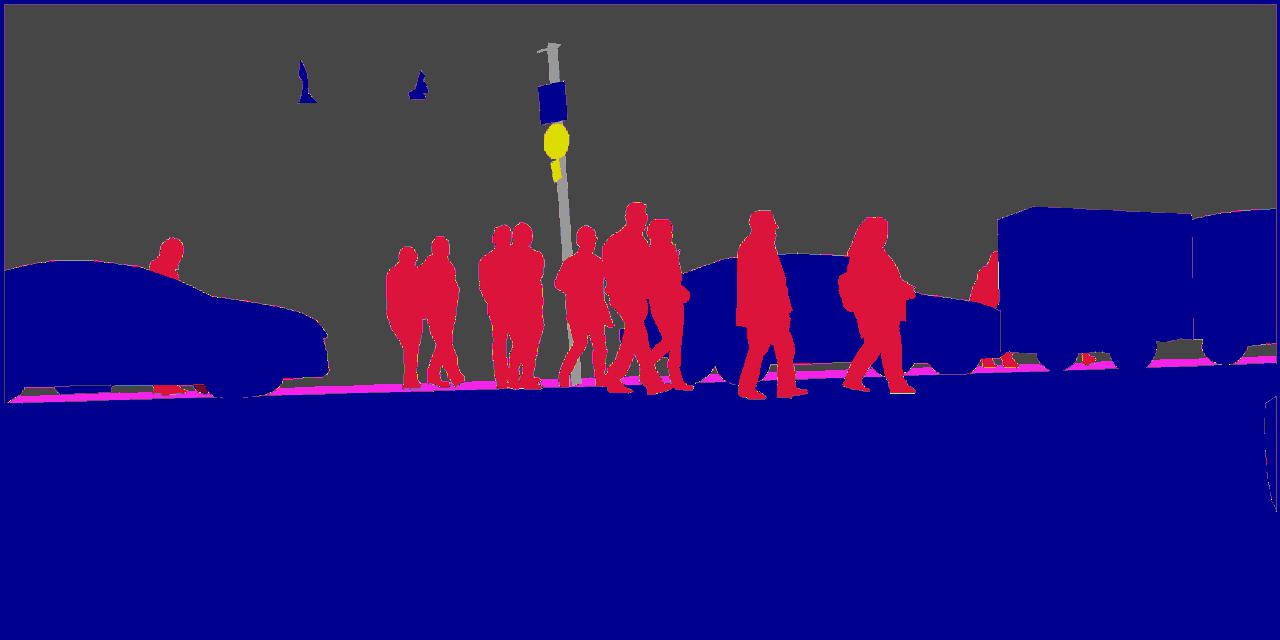} &\hspace{-0.47cm}
\includegraphics[width=0.182\linewidth, height=0.12\linewidth]{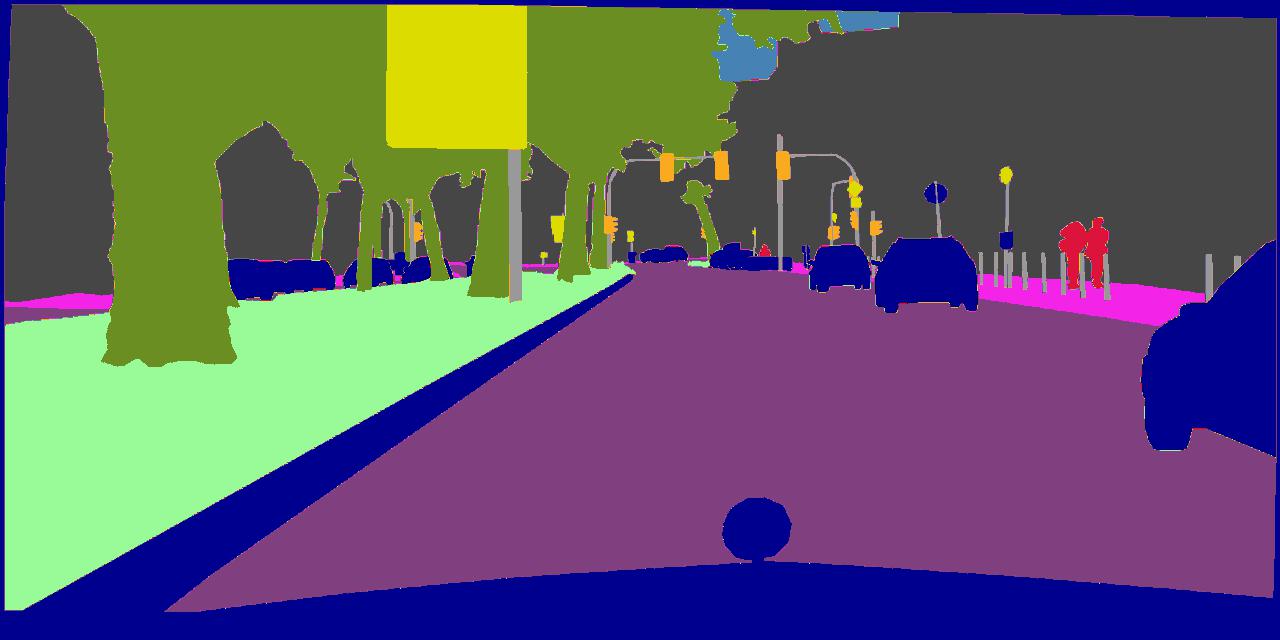} &\hspace{-0.47cm}
\includegraphics[width=0.182\linewidth, height=0.12\linewidth]{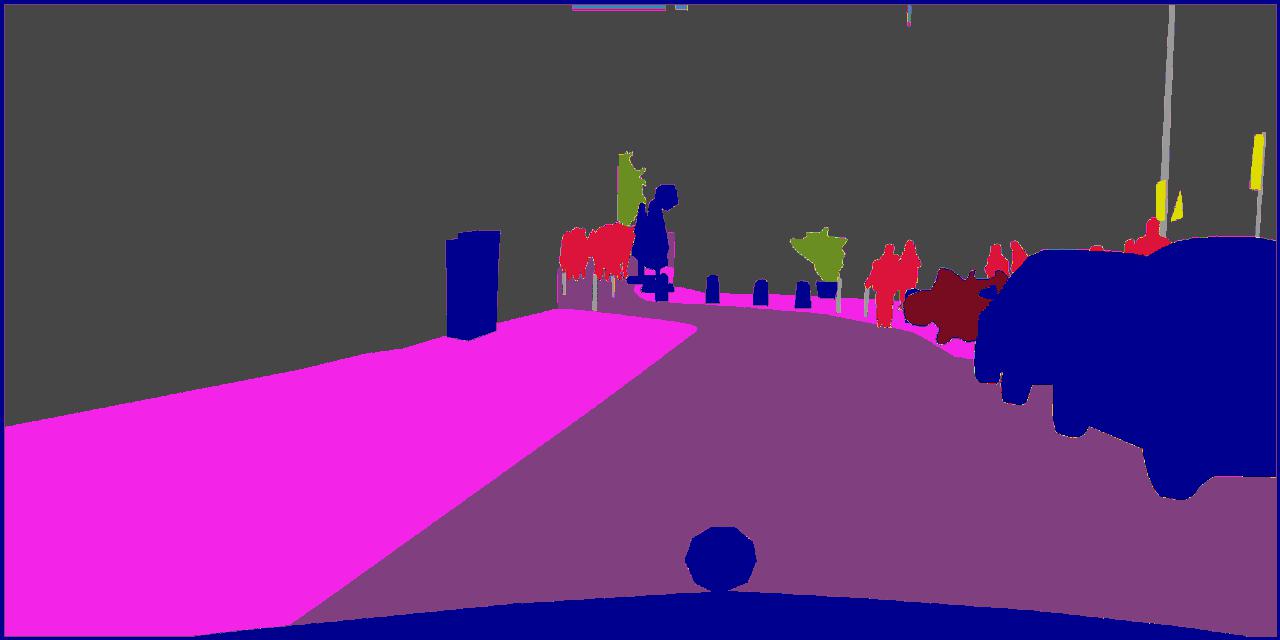} &\hspace{-0.47cm}
\includegraphics[width=0.182\linewidth, height=0.12\linewidth]{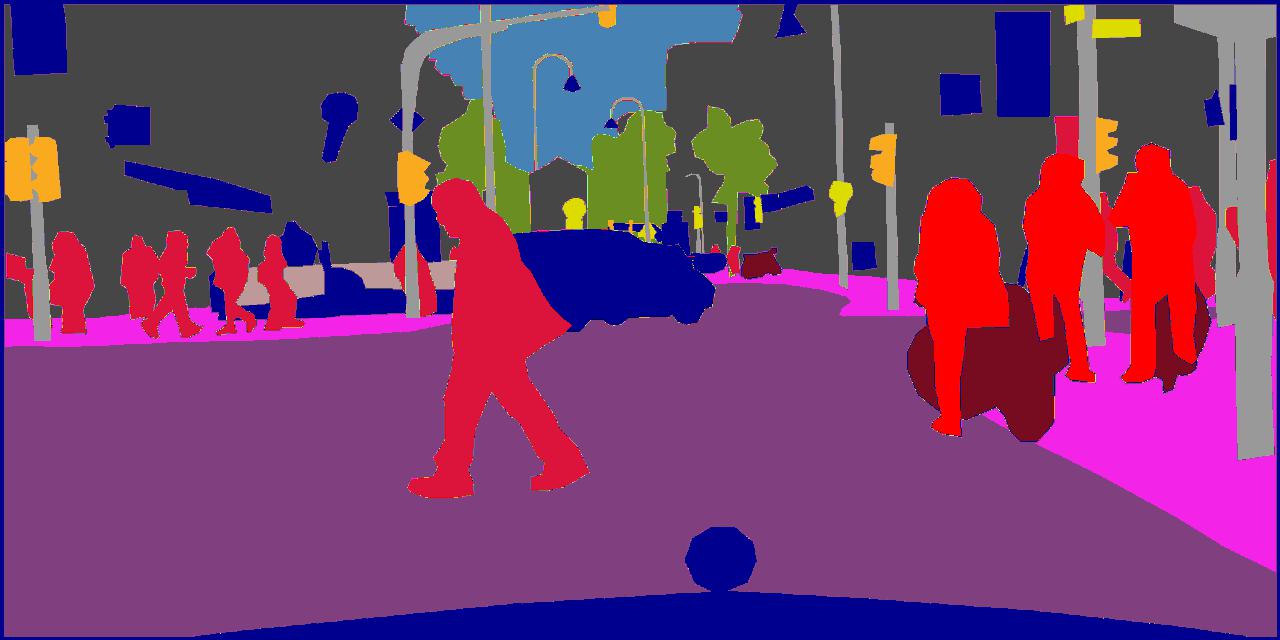} &\hspace{-0.47cm}
\includegraphics[width=0.188\linewidth, height=0.12\linewidth]{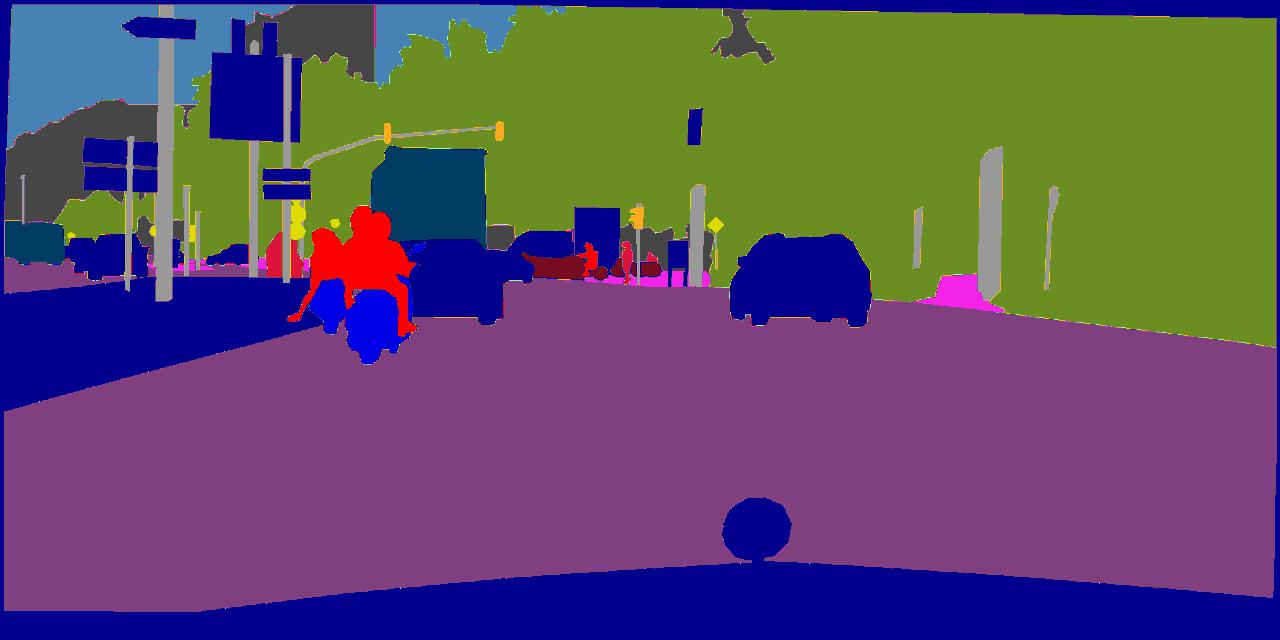} \\ \cline{1-2}
\multirow{2}{*}{\rotatebox[origin=c]{90}{\parbox[c]{0.8cm}{\textbf{Apolloscapes}}}} &\verticaltext[26pt]{\small}{\textbf{Raw Images}} &\hspace{-0.28cm}
\includegraphics[width=0.182\linewidth, height=0.12\linewidth]{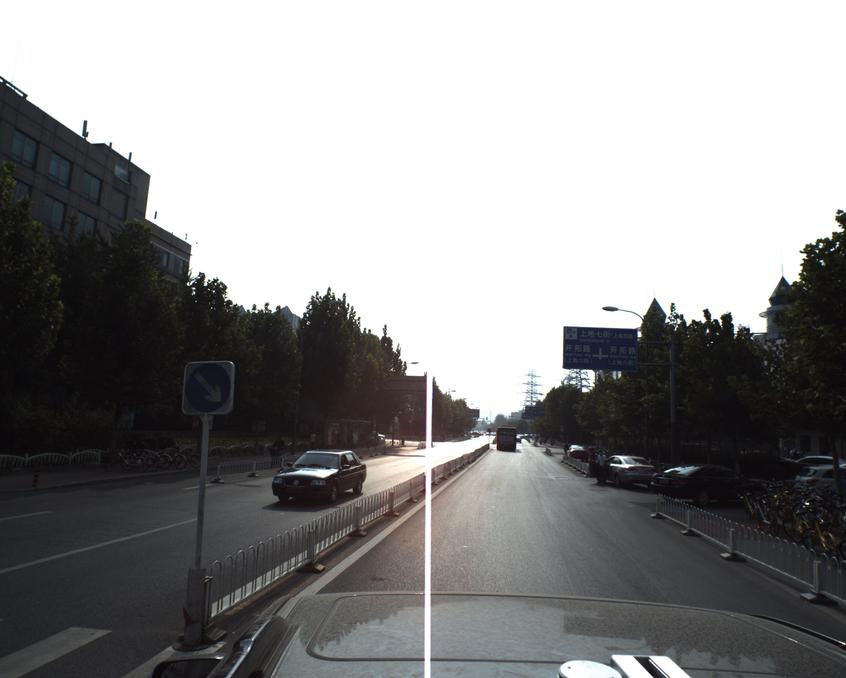} &\hspace{-0.47cm}
\includegraphics[width=0.182\linewidth, height=0.12\linewidth]{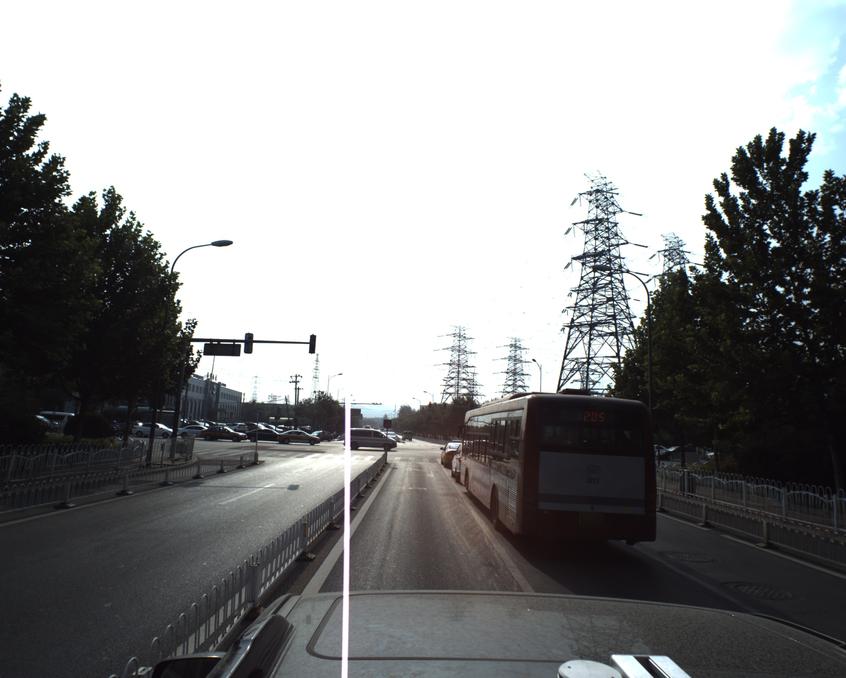} &\hspace{-0.47cm}
\includegraphics[width=0.182\linewidth, height=0.12\linewidth]{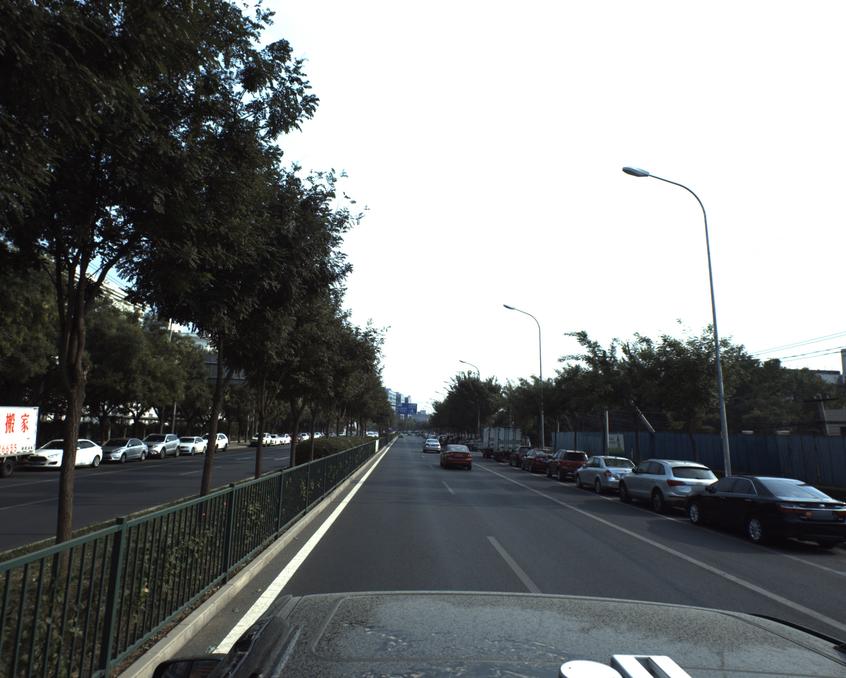} &\hspace{-0.47cm}
\includegraphics[width=0.182\linewidth, height=0.12\linewidth]{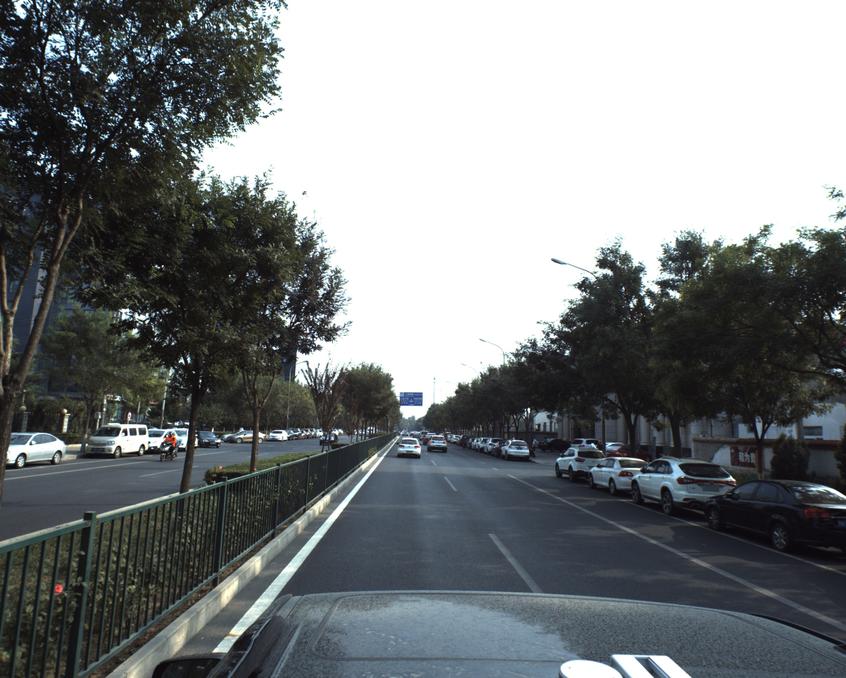} &\hspace{-0.47cm}
\includegraphics[width=0.188\linewidth, height=0.12\linewidth]{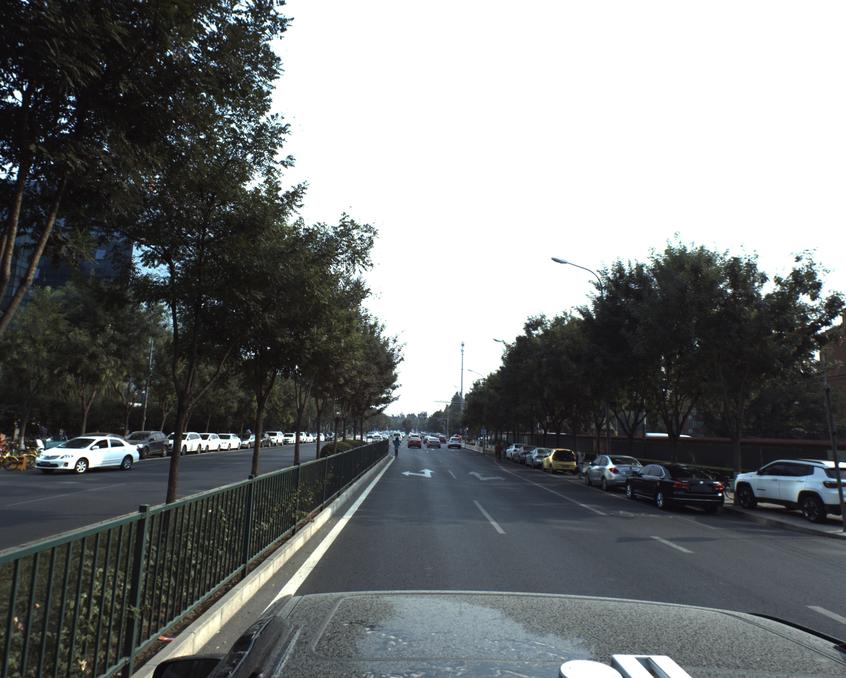}  \\ \cline{2-2}
                  &\verticaltext[26pt]{\small}{\textbf{Annotations}} &\hspace{-0.28cm}
\includegraphics[width=0.182\linewidth, height=0.12\linewidth]{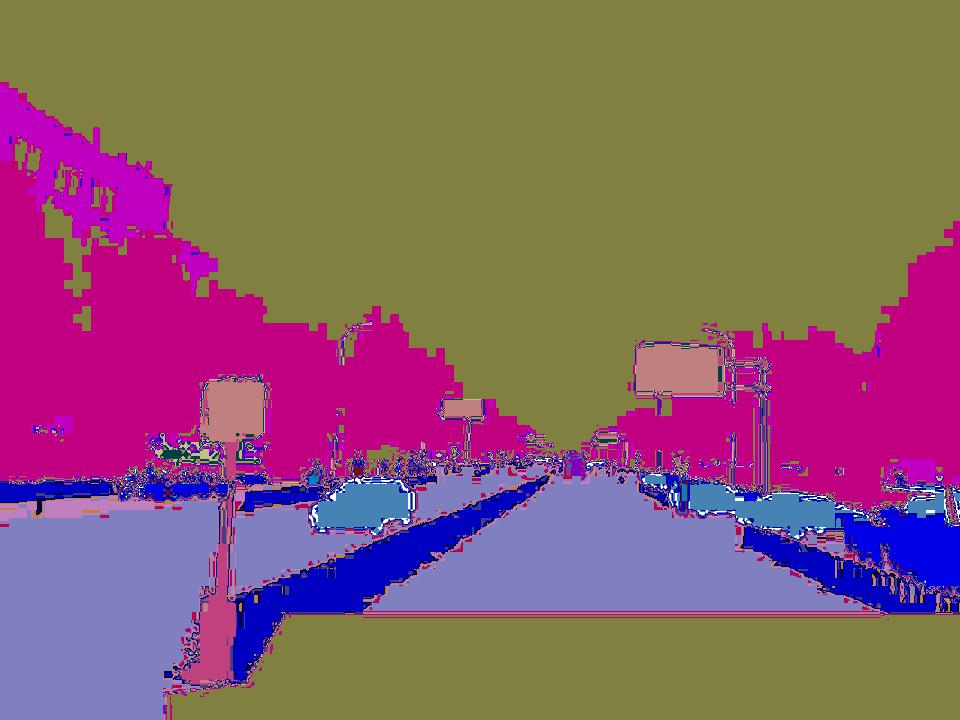} &\hspace{-0.47cm}
\includegraphics[width=0.182\linewidth, height=0.12\linewidth]{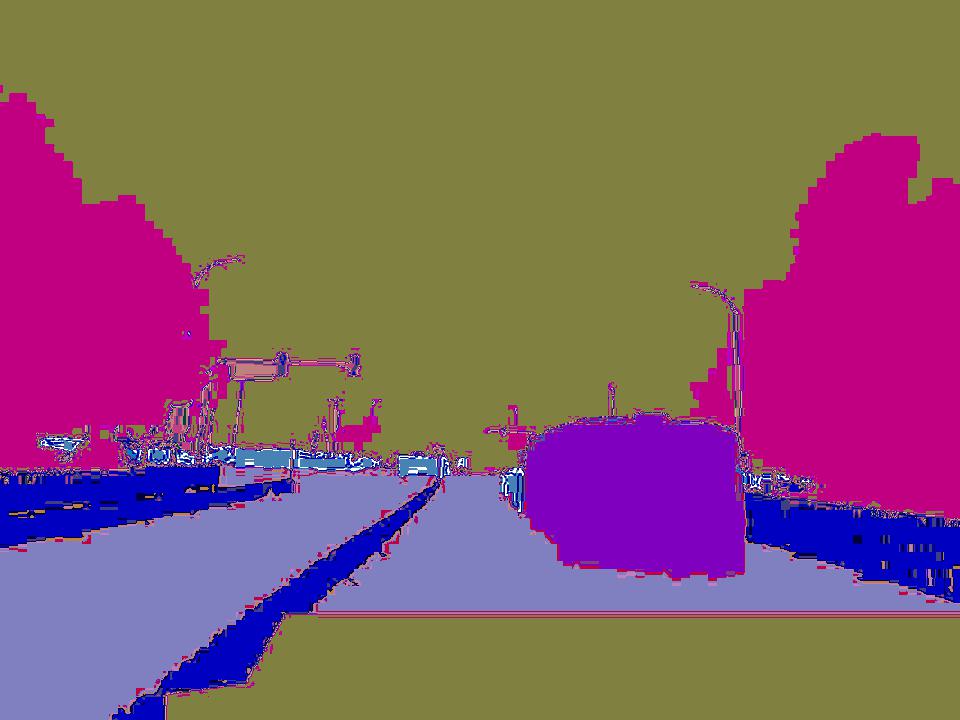} &\hspace{-0.47cm}
\includegraphics[width=0.182\linewidth, height=0.12\linewidth]{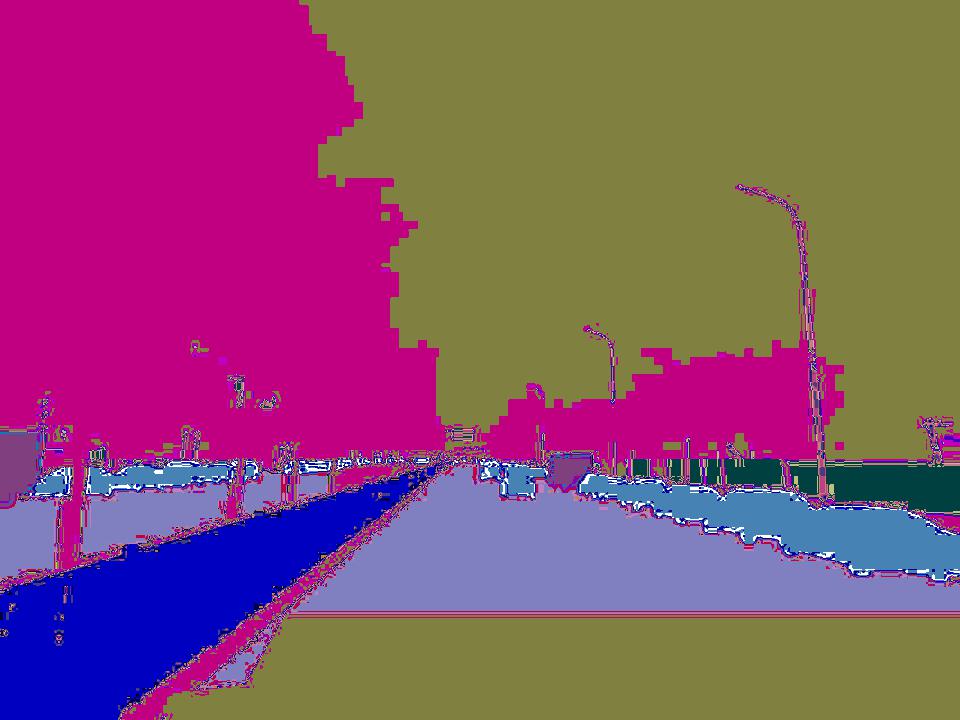} &\hspace{-0.47cm}
\includegraphics[width=0.182\linewidth, height=0.12\linewidth]{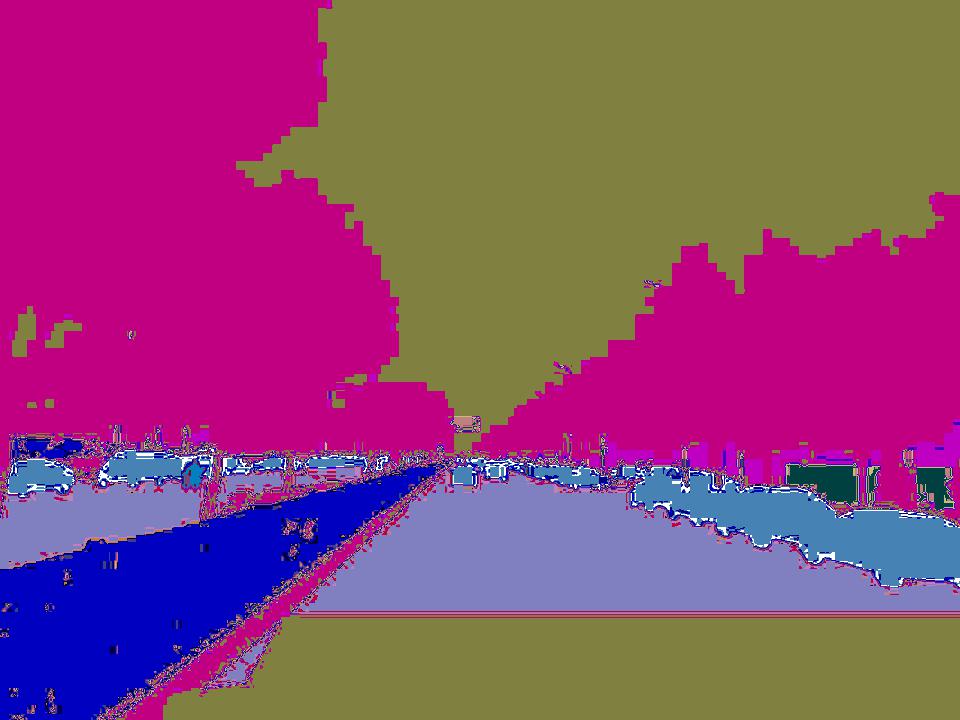} &\hspace{-0.47cm}
\includegraphics[width=0.188\linewidth, height=0.12\linewidth]{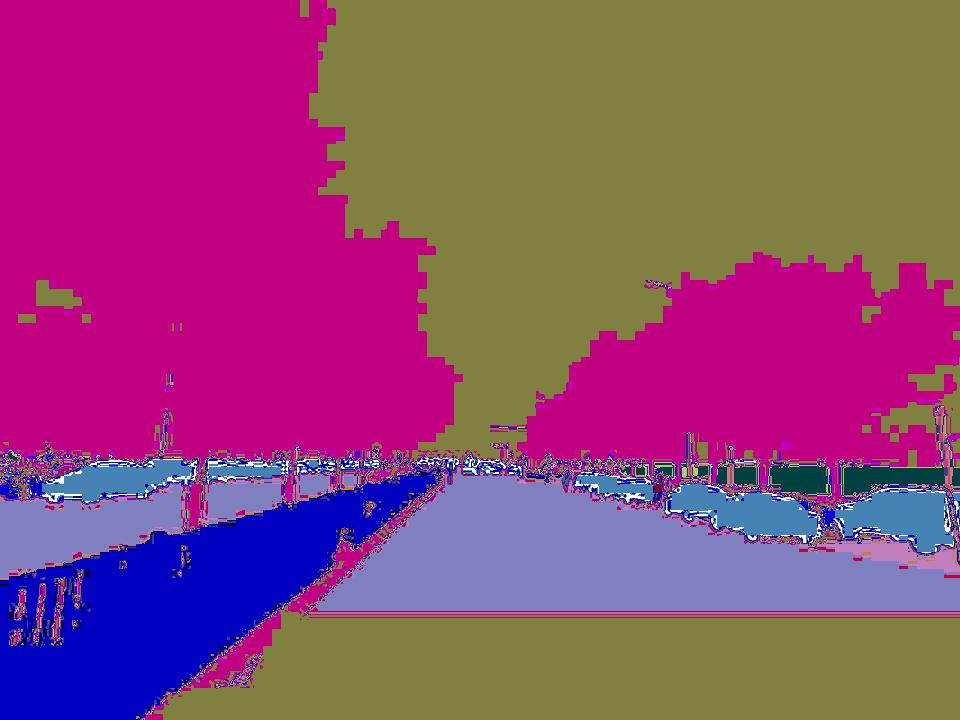}  \\ \cline{1-2}
\multirow{2}{*}{\rotatebox[origin=c]{90}{\parbox[c]{0.9cm}{\textbf{CARLA\_ADV}}}} &\verticaltext[26pt]{\small}{\textbf{Raw Images}} &\hspace{-0.28cm}
\includegraphics[width=0.182\linewidth, height=0.12\linewidth]{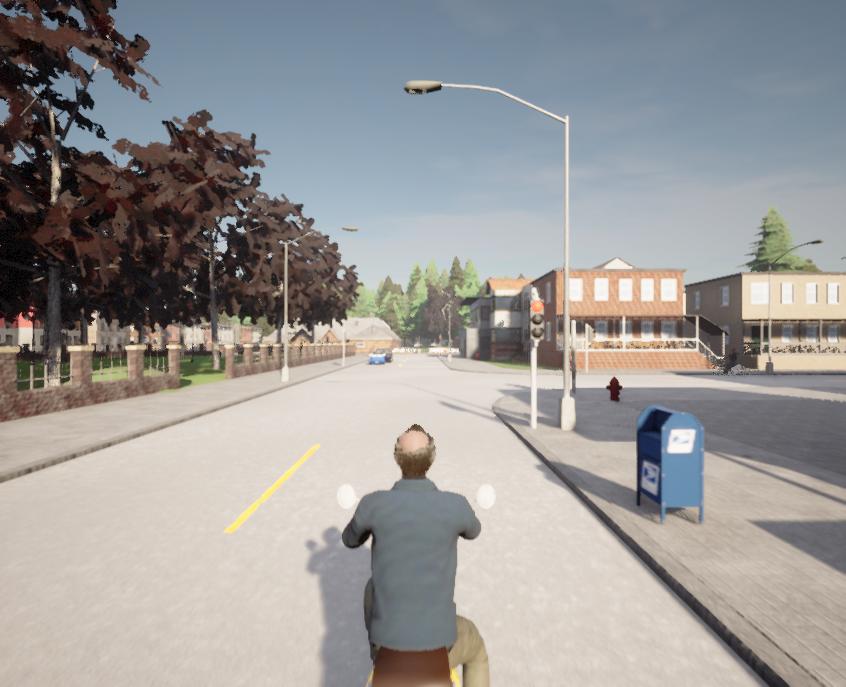} &\hspace{-0.47cm}
\includegraphics[width=0.182\linewidth, height=0.12\linewidth]{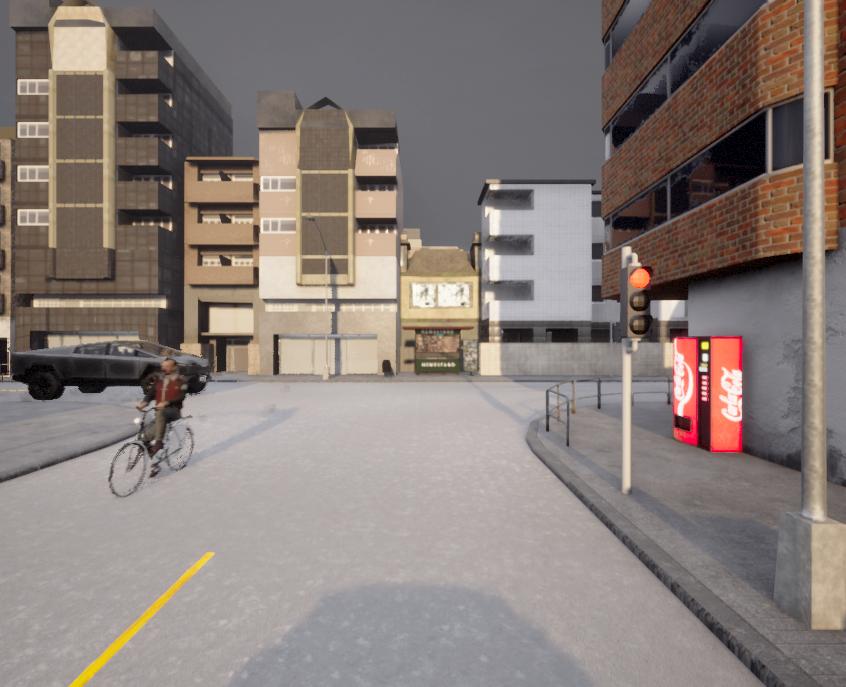} &\hspace{-0.47cm}
\includegraphics[width=0.182\linewidth, height=0.12\linewidth]{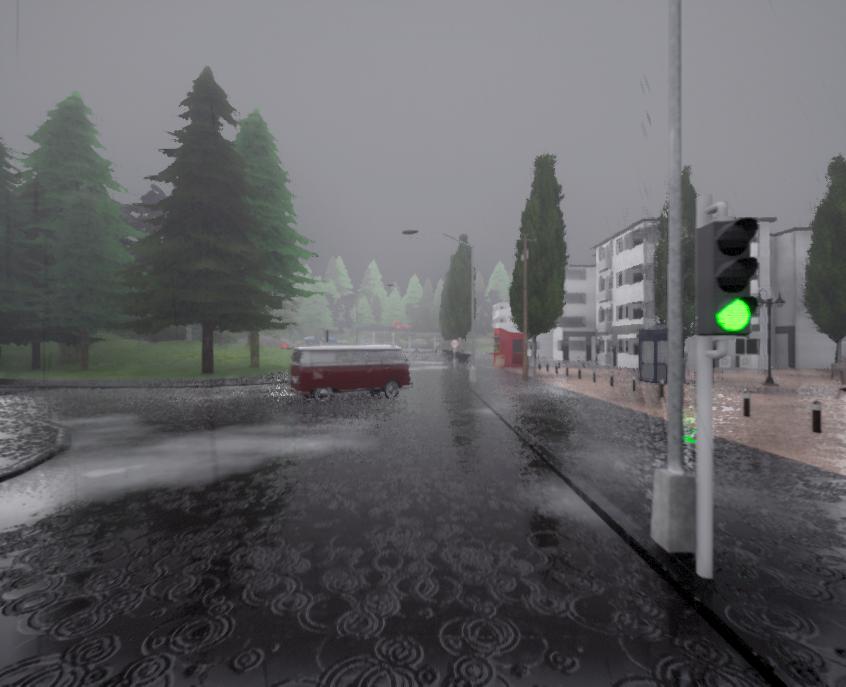} &\hspace{-0.47cm}
\includegraphics[width=0.182\linewidth, height=0.12\linewidth]{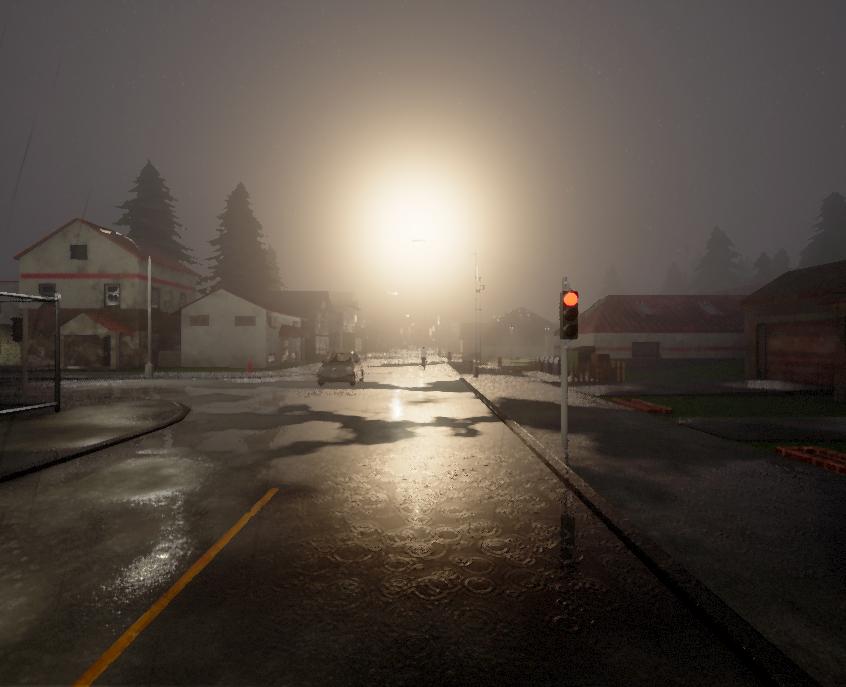} &\hspace{-0.47cm}
\includegraphics[width=0.188\linewidth, height=0.12\linewidth]{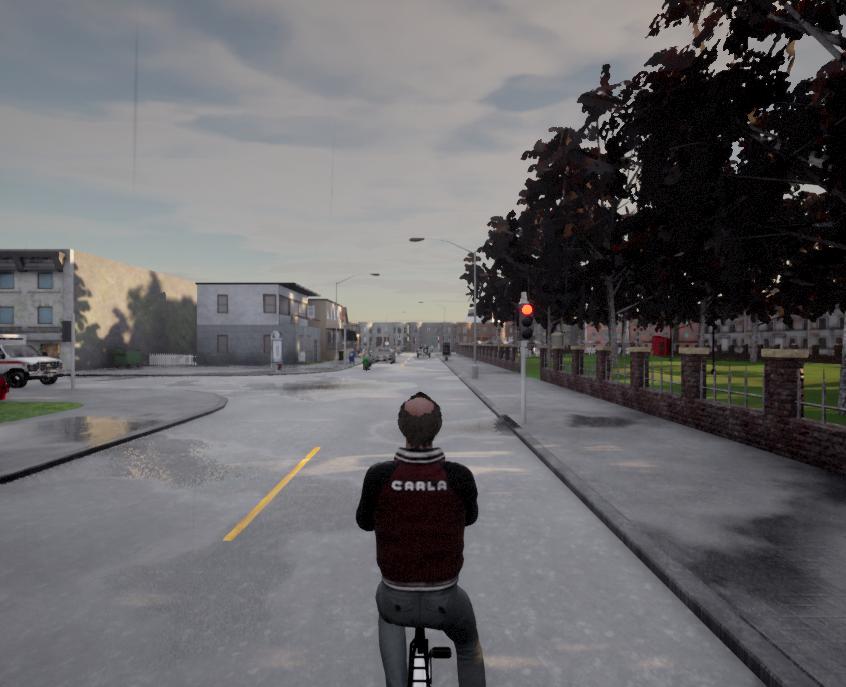}  \\ \cline{2-2}
                  &\verticaltext[26pt]{\small}{\textbf{Annotations}} &\hspace{-0.28cm}
\includegraphics[width=0.182\linewidth, height=0.12\linewidth]{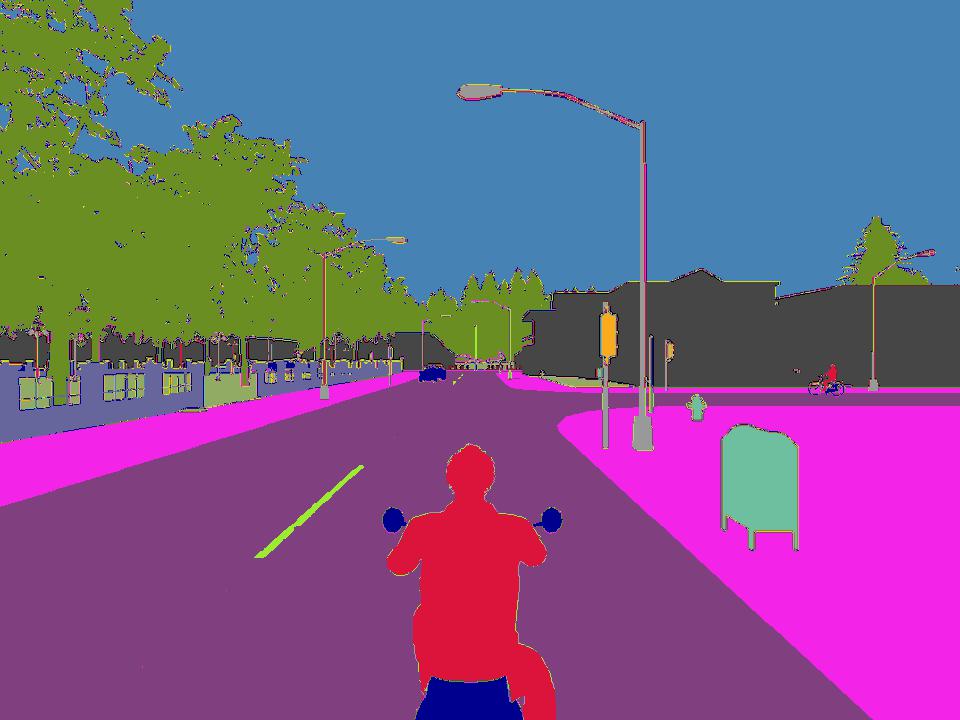} &\hspace{-0.47cm}
\includegraphics[width=0.182\linewidth, height=0.12\linewidth]{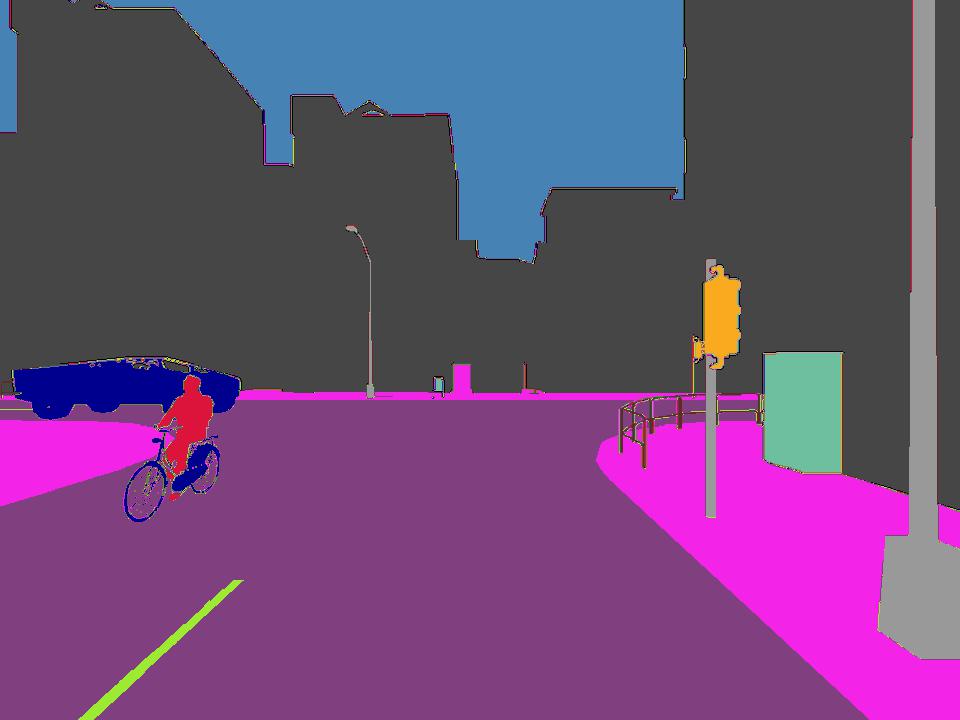} &\hspace{-0.47cm}
\includegraphics[width=0.182\linewidth, height=0.12\linewidth]{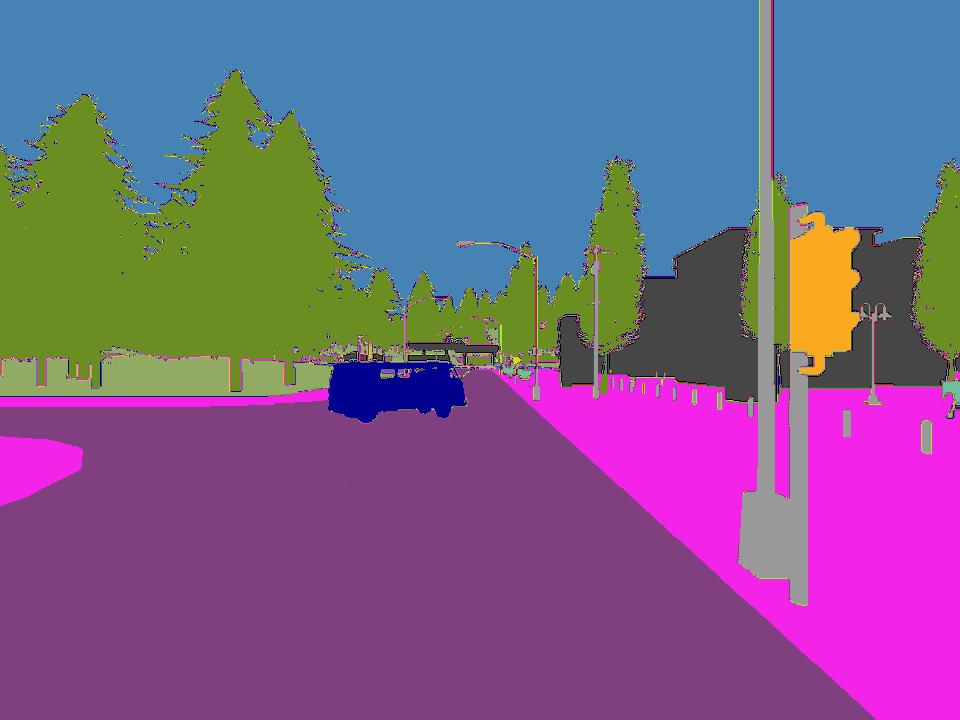} &\hspace{-0.47cm}
\includegraphics[width=0.182\linewidth, height=0.12\linewidth]{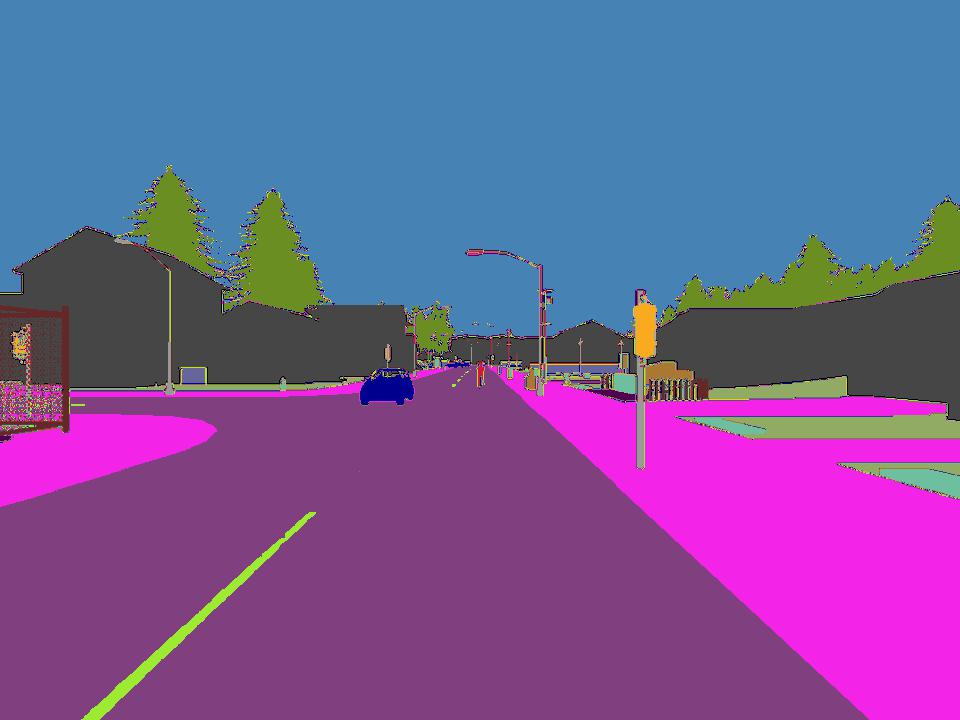} &\hspace{-0.47cm}
\includegraphics[width=0.188\linewidth, height=0.12\linewidth]{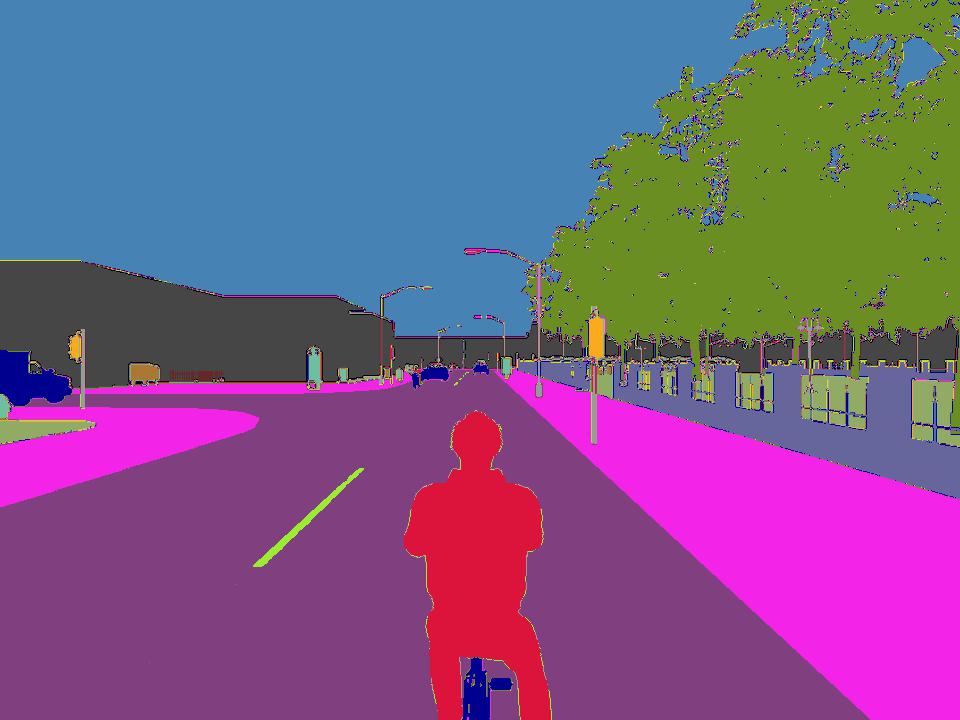}  \\ \hline
\end{tabularx}
\label{Tab:qua_perf_LAM}
\vspace{-0.1cm}
\end{table*}

\subsubsection{Qualitative Evaluation}
\begin{figure*}[tp]
    \centering
    \begin{minipage}{0.2\linewidth}
        \centering
        \subfloat[Raw Image]{
            \includegraphics[width=\linewidth,height=0.575\linewidth]{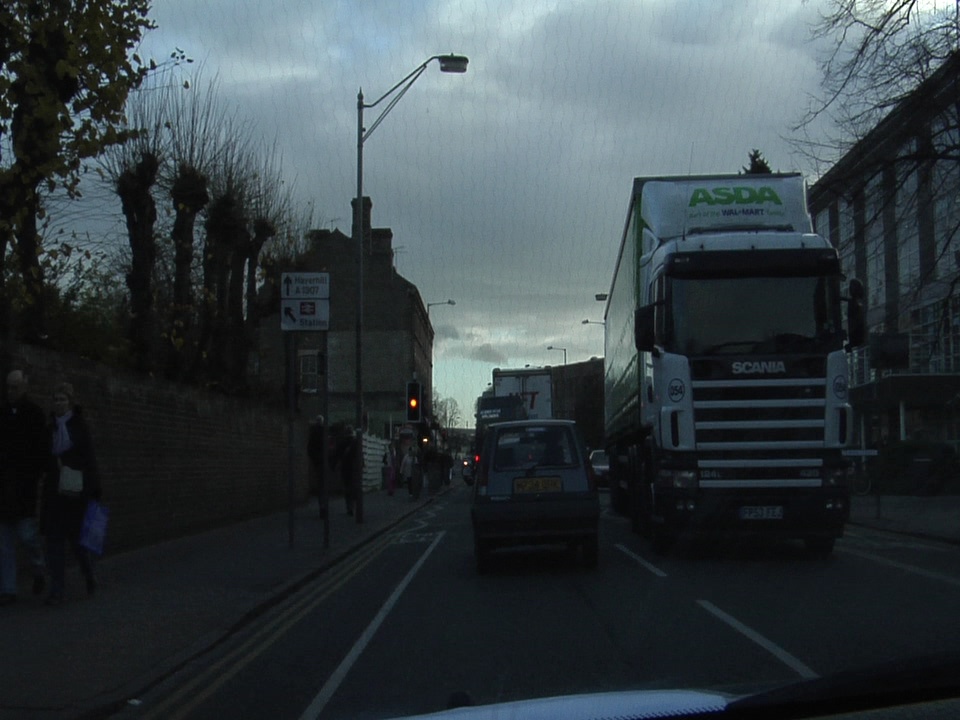}
            \label{fig:sub1}
        }
        \vspace{-0.28cm}
        \subfloat[LAM's Label]{
            \includegraphics[width=\linewidth,height=0.575\linewidth]{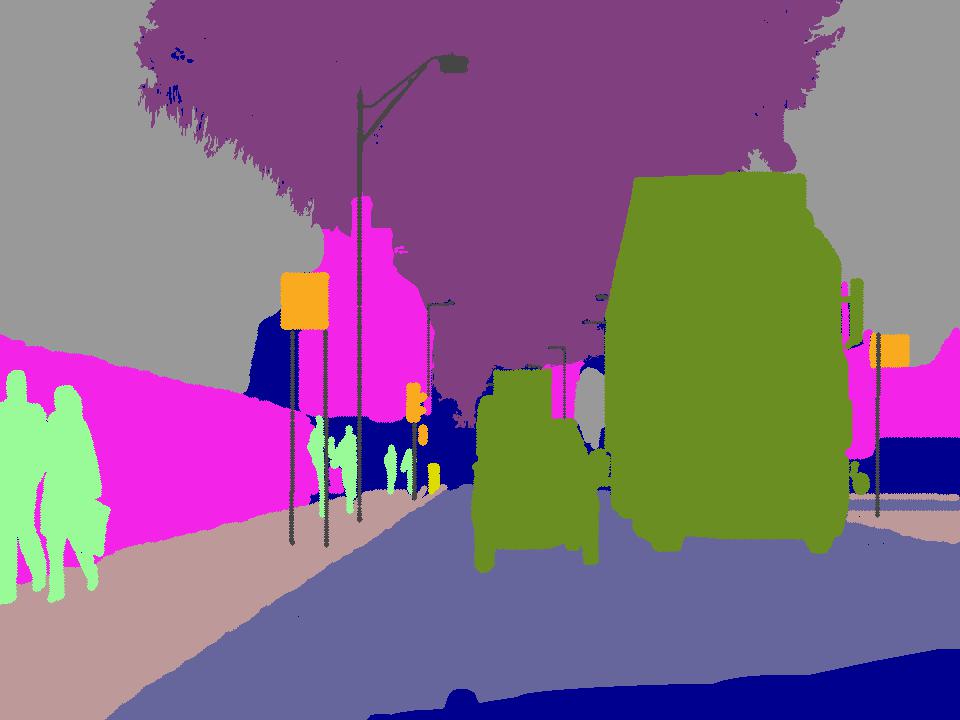}
            \label{fig:sub2}
        }
    \caption{Image \& Label}
    \label{Fig:indepth_example}
    \end{minipage}%
    \begin{minipage}{0.85\linewidth}
    \vspace{-0.3cm}
    \captionof{table}{Output comparison of OptOU against SCA}
    \vspace{0.3cm}
    \centering
    \begin{tabular}{|c|@{}c@{}@{}c@{}@{}c@{}@{}c@{}|}
    \hline
    \multirow{2}{*}{} & \multicolumn{1}{c|}{\multirow{2}{*}{SCA's Ouput}} & \multicolumn{3}{c|}{OptOU's Output}                                             \\ \cline{3-5} 
                      & \multicolumn{1}{c|}{}                             & \multicolumn{1}{c|}{1-th Layer} & \multicolumn{1}{c|}{5-th Layer} & 10-th Layer \\ \hline
   \rotatebox[origin=c]{90}{\parbox[c]{2.0cm}{\small\textbf{Chan\#2~(Pole)}}}  &
   \parbox[c]{0.21\linewidth}{\centering\includegraphics[width=0.99\linewidth, height=0.63\linewidth]{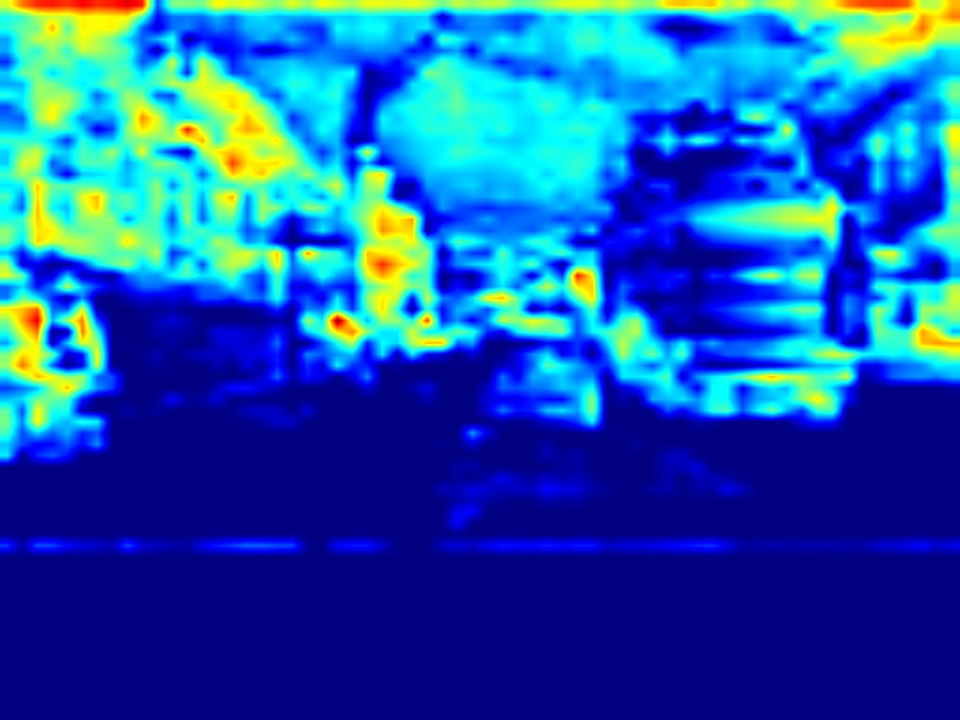}} &
   \parbox[c]{0.21\linewidth}{\centering\includegraphics[width=0.99\linewidth, height=0.65\linewidth]{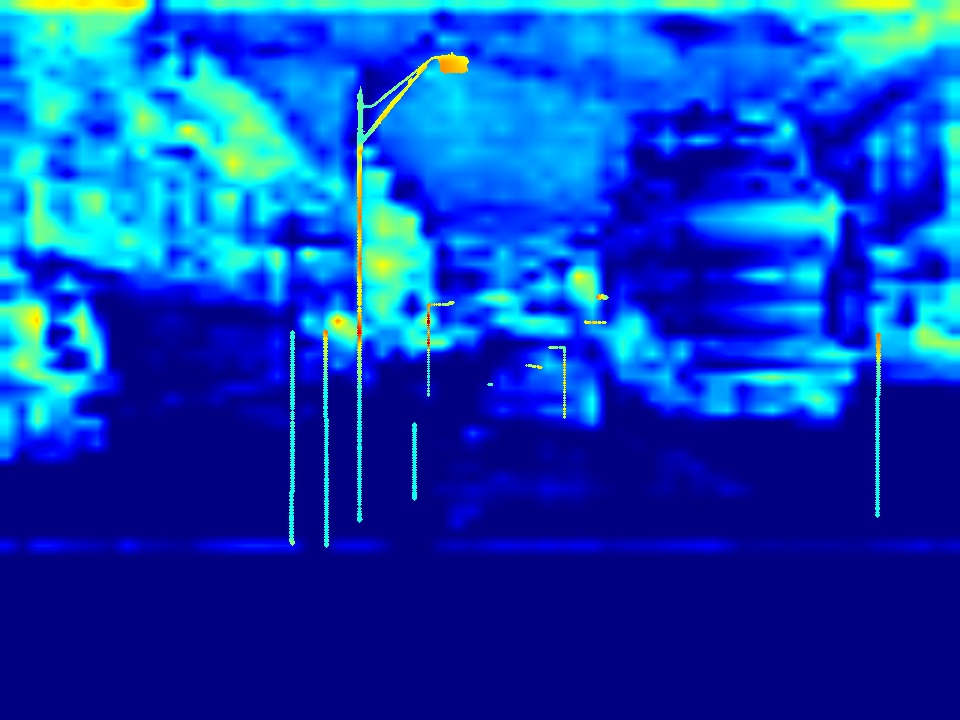}} &
   \parbox[c]{0.21\linewidth}{\centering\includegraphics[width=0.99\linewidth, height=0.63\linewidth]{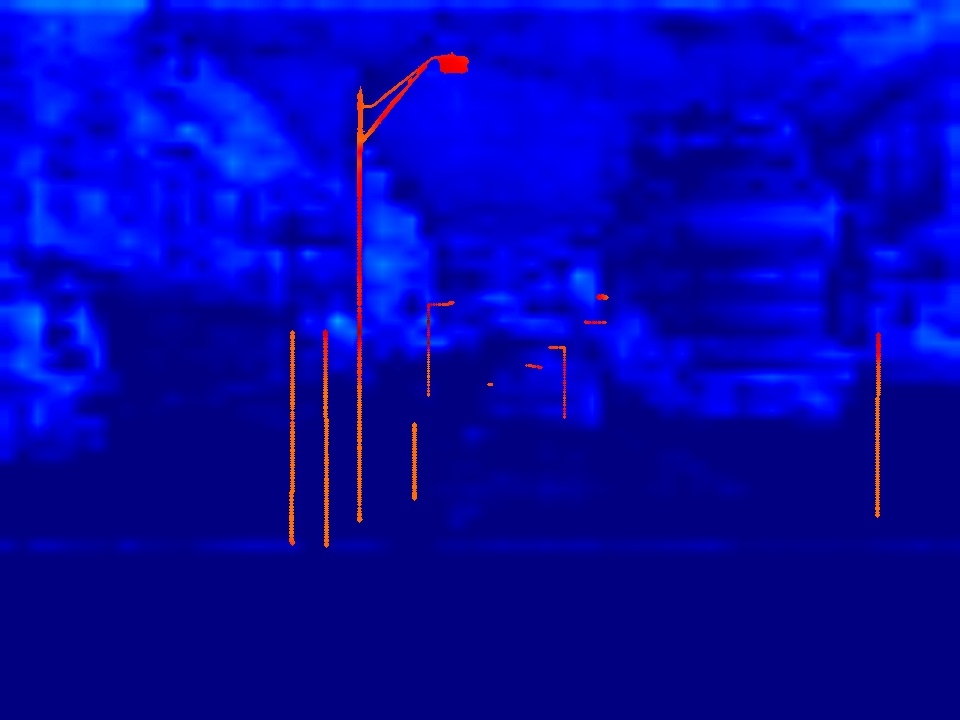}} &
   \parbox[c]{0.21\linewidth}{\centering\includegraphics[width=0.99\linewidth, height=0.65\linewidth]{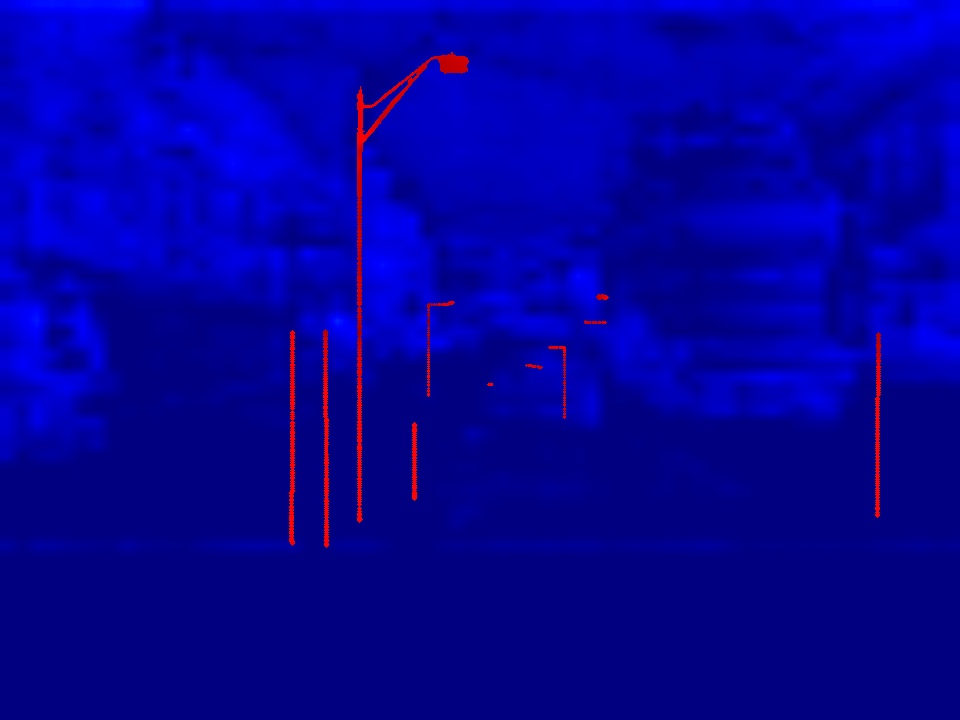}}   \\ \cline{1-1}
   \rotatebox[origin=c]{90}{\parbox[c]{2.0cm}{\small\textbf{Chan\#8~(Car)}}} &
   \parbox[c]{0.21\linewidth}{\centering\includegraphics[width=0.99\linewidth, height=0.65\linewidth]{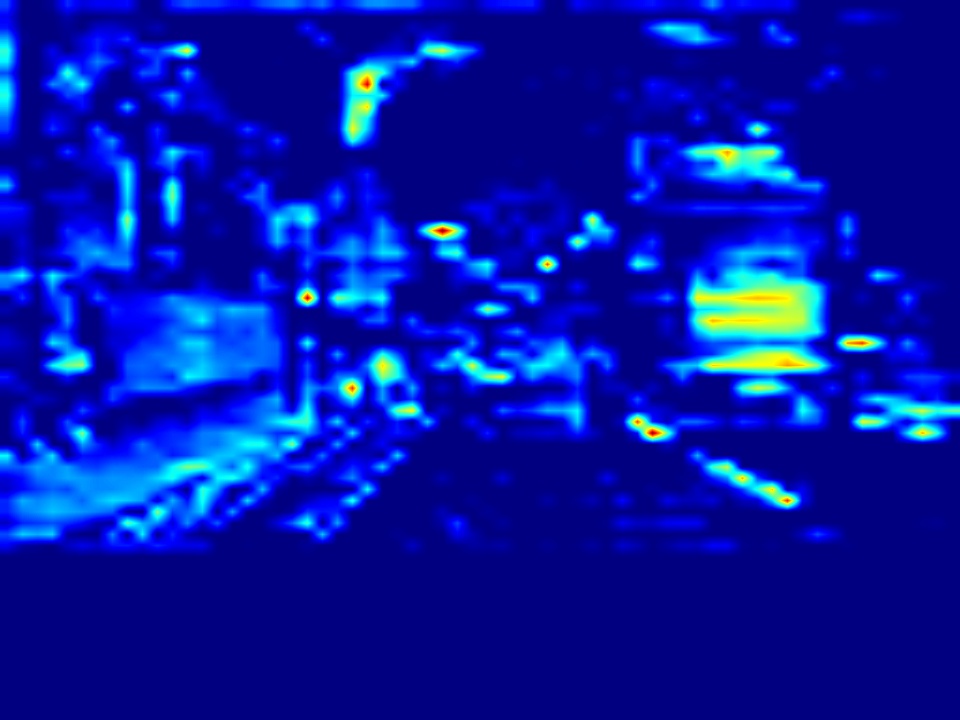}} &
   \parbox[c]{0.21\linewidth}{\centering\includegraphics[width=0.99\linewidth, height=0.63\linewidth]{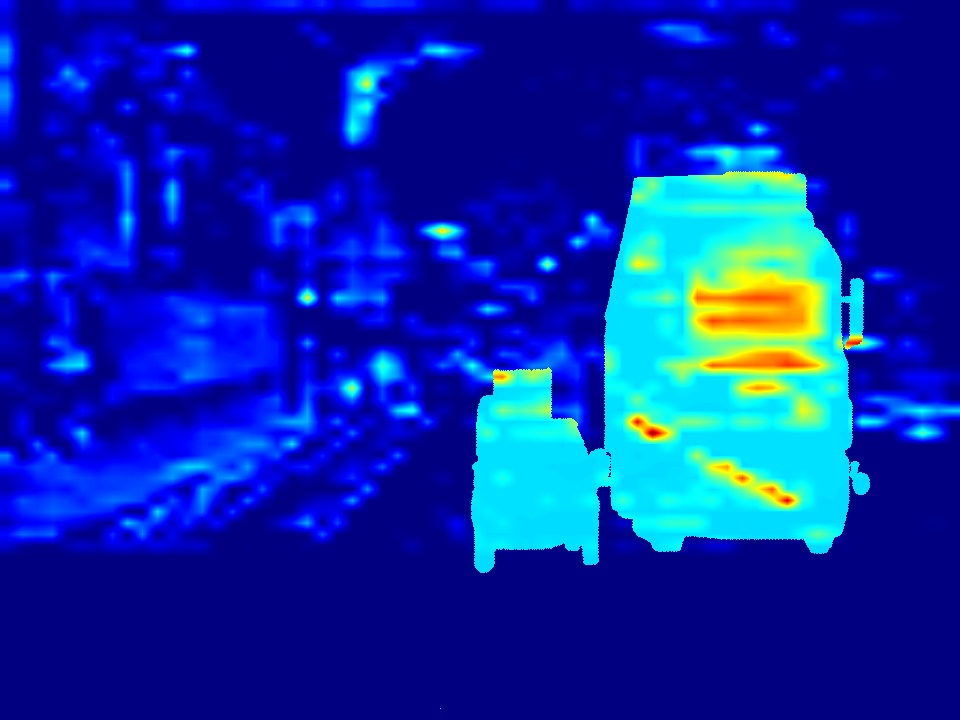}} &
   \parbox[c]{0.21\linewidth}{\centering\includegraphics[width=0.99\linewidth, height=0.65\linewidth]{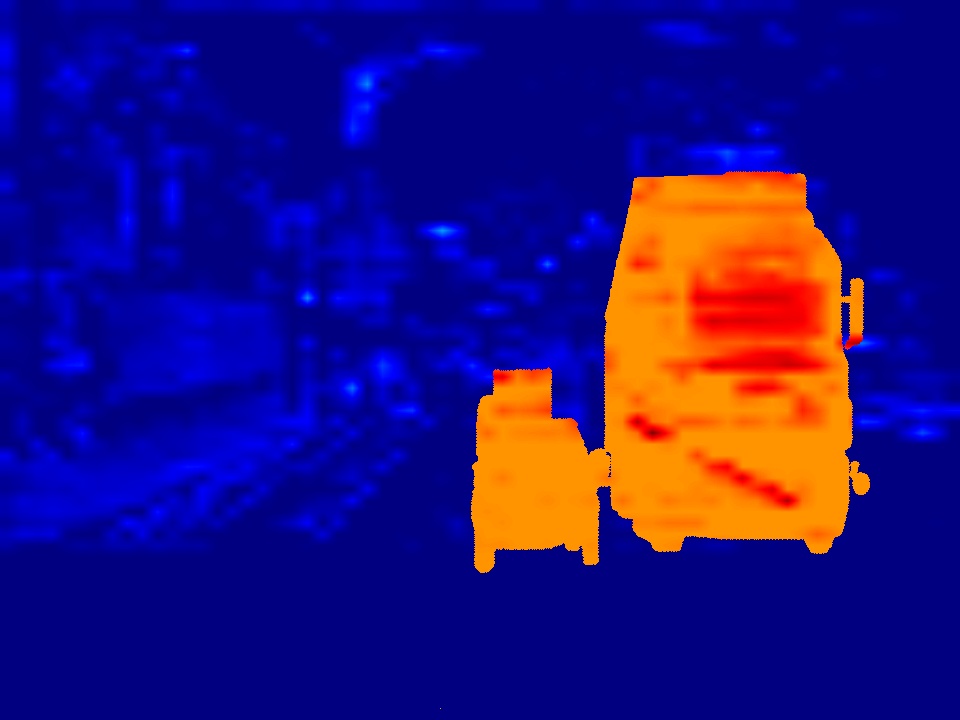}} &
   \parbox[c]{0.21\linewidth}{\centering\includegraphics[width=0.99\linewidth, height=0.63\linewidth]{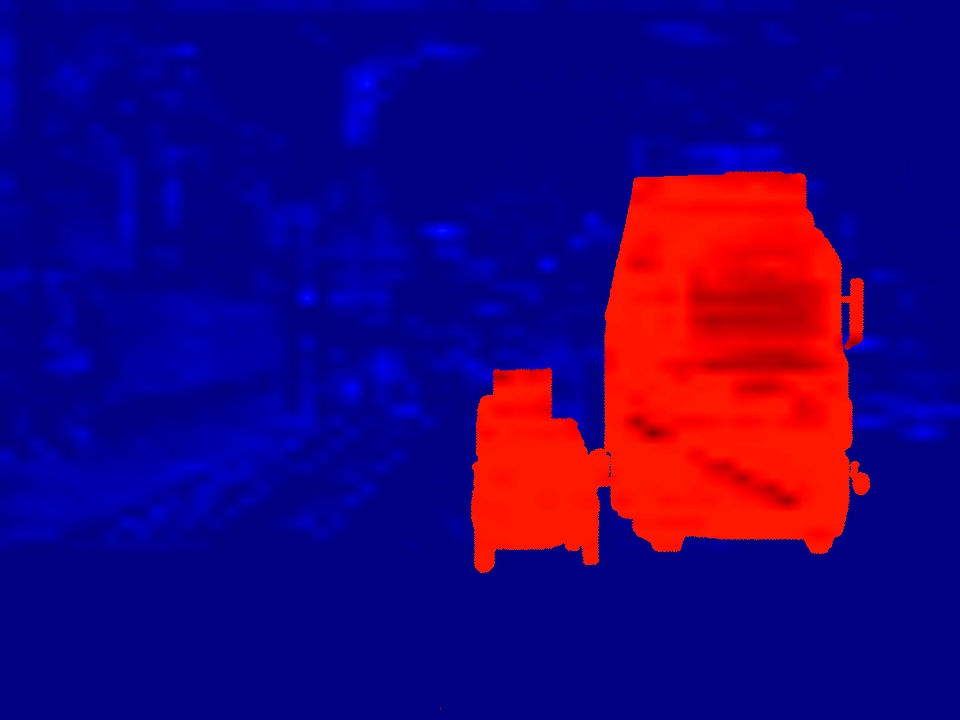}} \\ \hline
    \end{tabular}
    \label{Tab:indepth_exploration}
\end{minipage} 
\end{figure*}

\Cref{Tab:qua_perf_LAM} presents a qualitative comparison of LAM's performance across CamVid, Cityscapes, Apolloscapes, and CARLA\_ADV datasets. Each dataset showcases five distinct examples. \Cref{Tab:qua_perf_LAM} visually demonstrates LAM's annotation capabilities across diverse driving scenarios. Specifically, LAM's annotations are class-aware (for example, trees are assigned to identical semantic IDs (colors)) and accurate annotation (for instance, tree leaves are annotated quite fine).

\subsection{In-depth Exploration of OptOU}
This section details more in-depth investigation about the proposed OptOU. \Cref{Fig:indepth_example} presents a CamVid's RGB image and the LAM-generated label. \Cref{Tab:indepth_exploration} compares the output of SCA and OptOU, where the output contains multiple channels and each channel represents one semantic class. For the output of the image in \Cref{Fig:indepth_example}, channel 2 represents the class of \textbf{Pole} while channel 8 represents the class of \textbf{Car}. We can observe following patterns for both channels: (I) SCA's output directly mapping from ViT is not capable of segmenting objects (\ie, \textbf{Pole} and \textbf{Car}) from the image. (II) OptOU's output is more clear in segmenting objects than that of SCA. Concretely, each layer's output of OptOU segments objects clearly whereas SCA's output does not. (III) The segmentation capabilities of OptOU are enhanced continuously as the optimization progresses. Specifically, the output of first layer, though, can segment the objects, it still contains much noises; in contrast, the output of $10$-th layer can segment objects more clearly with little noises. 

\section{Conclusion}
This paper introduces an interpretable, high-fidelity, and prompt-free annotator LAM. It consists of a pretrained ViT, the proposed SCA and OptOU. LAM just requires only one training RGB image owing to its quite small volume of learnable parameters. We carry out comprehensive experiments on real-world datasets and CARLA simulation dataset to verify LAM's labeling performance. Experimental results demonstrate that the proposed LAM can definitely generate finer labels. Future work plans to extend the proposed LAM to label other modalities, such as Lidar, depth camera, etc.

\end{document}